\pdfoutput=1
\documentclass[11pt]{article}

\PassOptionsToPackage{table,dvipsnames}{xcolor}

\usepackage[final]{acl}

\usepackage{times}
\usepackage{latexsym}

\usepackage[T5]{fontenc}
\usepackage[utf8]{inputenc}

\usepackage{microtype}

\usepackage{inconsolata}

\usepackage{graphicx}

\usepackage{multirow}
\usepackage{adjustbox} 
\usepackage{booktabs}
\usepackage{soul}
\usepackage{threeparttable}
\usepackage{multirow}
\usepackage{makecell}
\usepackage{arydshln} % dashed cmidrule
\usepackage{listings}
\usepackage{pgf}

\newcommand{\greygra}[1]{%
  \pgfmathparse{int(0 + (#1 * 60))}%  
  \edef\percent{\pgfmathresult}%
  \xdef\mycolor{black!\percent}%
  \cellcolor{\mycolor}#1
}

\title{WETBench: A Benchmark for Detecting Task-Specific Machine-Generated Text on Wikipedia}

\author{%
  Gerrit Quaremba$^{1}$, Elizabeth Black$^{1}$, Denny Vrandečić$^{2}$, Elena Simperl$^{1}$ \\
  $^{1}$King's College London, 
  $^{2}$Wikimedia Foundation \\
  \texttt{\{gerrit.quaremba,elizabeth.black,elena.simperl\}@kcl.ac.uk} \\\texttt{denny@wikimedia.org}
}

\begin{document}
\maketitle

\begin{abstract}
Given Wikipedia's role as a trusted source of high-quality, reliable content, concerns are growing about the proliferation of low-quality machine-generated text (MGT) produced by large language models (LLMs) on its platform.  
Reliable detection of MGT is therefore essential. However, existing work primarily evaluates MGT detectors on generic generation tasks rather than on tasks more commonly performed by Wikipedia editors.  
This misalignment can lead to poor generalisability when applied in real-world Wikipedia contexts.  
We introduce \textbf{WETBench}, a multilingual, multi-generator, and \textit{task-specific} benchmark for MGT detection. We define three editing tasks, empirically grounded in Wikipedia editors’ perceived use cases for LLM-assisted editing: \textit{Paragraph Writing}, \textit{Summarisation}, and \textit{Text Style Transfer}, which we implement using two new datasets across three languages.  
For each writing task, we evaluate three prompts, generate MGT across multiple generators using the best-performing prompt, and benchmark diverse detectors.  
We find that, across settings, training-based detectors achieve an average accuracy of 78\%, while zero-shot detectors average 58\%.  
These results show that detectors struggle with MGT in realistic generation scenarios and underscore the importance of evaluating such models on diverse, task-specific data to assess their reliability in editor-driven contexts.\footnote{An extended version of this workshop paper appears in \citet{quaremba2026tsm}.}

\end{abstract}

\section{Introduction}
\label{sec:intro}

\begin{figure}[h]
    \centering
    \includegraphics[scale=.62]{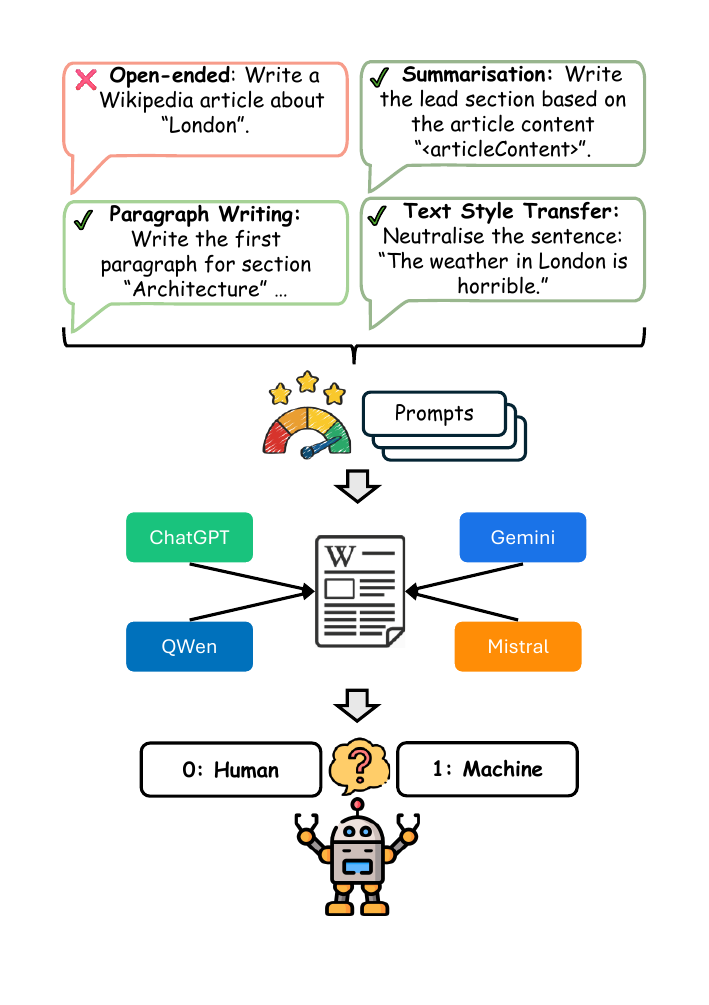}
    \vspace{-1em}
    \caption{We define \textit{\textcolor{ForestGreen}{task-specific editing scenarios}} on Wikipedia, test various prompting techniques, generate LLM-written text using the best-performing prompts, and benchmark SOTA detectors on these data. This contrasts with prior work, which primarily focuses on a single, \textit{\textcolor{red}{open-ended}} generation task that only partially captures the real-world editorial use of LLMs.}
    \label{fig:intro}
\end{figure}

Wikipedia serves as a vital source of high-quality, trustworthy data across artificial intelligence (AI) communities. Its scale and richness have played a foundational role in the development of large language models (LLMs)~\cite{deckelmann2023wikipedia,longpre2023}.  
However, the Wikipedia community has expressed growing concern about the increasing prevalence of machine-generated text (MGT) produced by LLMs on its platform.\footnote{\url{https://en.wikipedia.org/wiki/Wikipedia:Large_language_models}}  
The Wikimedia Foundation warns that the spread of low-quality, unreliable MGT in its projects could undermine its knowledge integrity.\footnote{\href{https://meta.wikimedia.org/w/index.php?title=Wikimedia_Foundation_Annual_Plan/2023-2024/Draft/External_Trends/Community_call_notes&oldid=24785109}{Wikipedia Community Call Notes 2023–24}}  
Specifically, unverified MGT poses challenges such as factual fabrication~\cite{huang2025survey} and the perpetuation of biases present in training data~\cite{gallegos2024}, both of which jeopardise Wikipedia's core content policies.\footnote{\url{https://en.wikipedia.org/wiki/Wikipedia:Core_content_policies}\label{wiki:pol}}  
Additionally, given Wikipedia's frequent inclusion in LLM training corpora, undetected MGT on the platform may contribute to performance degradation in future models~\cite{shumailov2024ai}.  
Consequently, distinguishing human-written from machine-generated text has become increasingly important, leading to community efforts to identify and remove MGT,\footnote{\url{https://en.wikipedia.org/wiki/Wikipedia:WikiProject_AI_Cleanup}} and to a growing body of research on estimating the prevalence of MGT on Wikipedia~\cite{brooks2024,huang2025wikipedia}.

Prior work on benchmarking MGT detectors~\cite[e.g.,][]{guo2023closechatgpthumanexperts,li2023mage,wang2023m4,wang2024m4gt} has included the Wikipedia domain but typically fails to reflect the complexities of editor-driven MGT instances on the platform.  
Existing experimental setups generally assume that MGT on Wikipedia results from (\textit{i}) open-ended, topic-to-text generation and (\textit{ii}) simplistic prompting techniques.  
These setups usually rely on a single prompt to generate an entire article, which diverges significantly from real-world Wikipedia editing practices that are task-specific and incremental.  
In fact, prompting an LLM to verbatim \textit{"Write a Wikipedia article about [...]}," as done in earlier work, is explicitly discouraged by the community.\footnote{\url{https://en.wikipedia.org/wiki/Wikipedia:Large_language_models}}

\begin{figure*}[h]
    \centering
    \includegraphics[scale=.3]{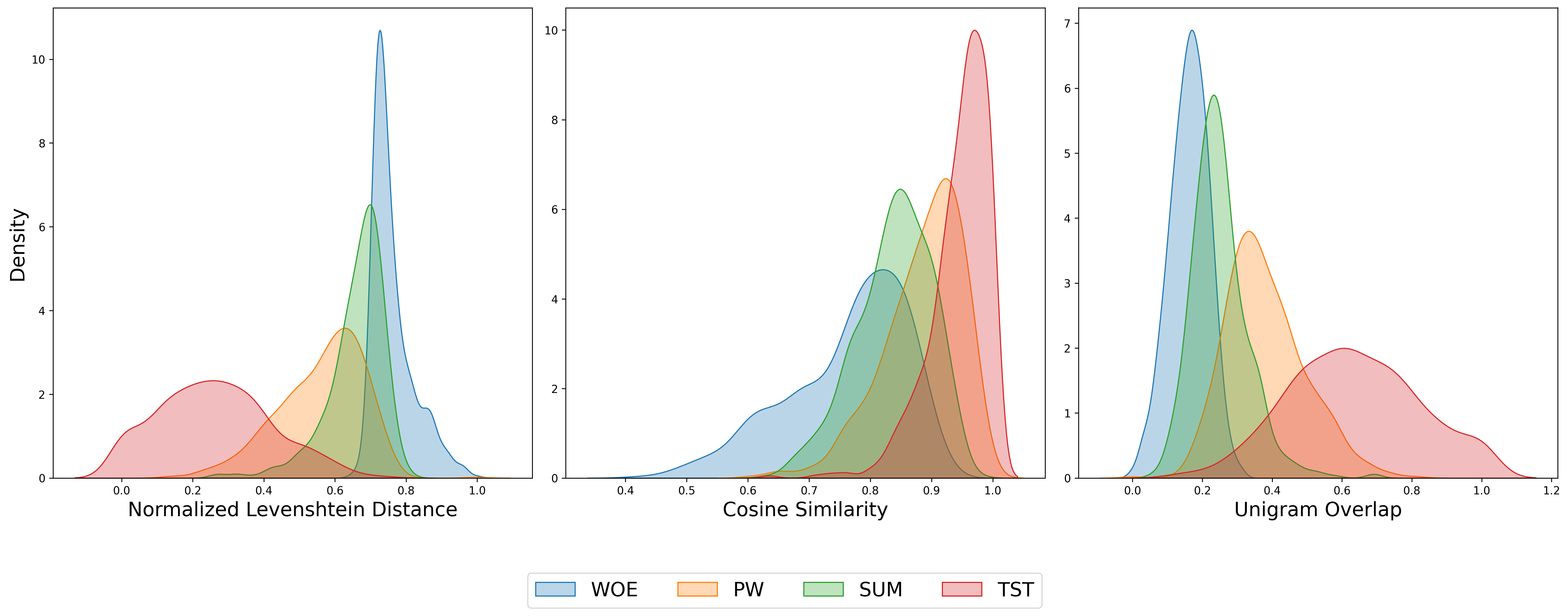}
    \caption{Comparison of MGT and HWT (N=600) for English Wikipedia Open-Ended Generation (WOE) vs. our Wikipedia editing tasks: Paragraph Writing (PW), Summarisation (SUM), and Text Style Transfer (TST). Task-specific MGT consistently demonstrates closer proximity to human writing across all dimensions.}
    \label{fig:dist}
\end{figure*}

These limitations in existing setups may obscure the actual performance of state-of-the-art (SOTA) detectors when applied to real-world Wikipedia contexts.  
Figure~\ref{fig:dist} shows that the textual characteristics of task-specific MGT—unlike open-ended, topic-to-text MGT—more closely resemble their human-written text (HWT) references.  
Detectors trained and evaluated on generic generation tasks may learn high-level textual patterns that are less transferable to task-specific MGT instances.  
Consequently, detectors may not generalise well to detecting diverse, task-specific MGT on Wikipedia, leaving an unknown number of instances with potentially harmful characteristics—such as hallucination or bias—largely undetected.  
To address this issue, we advocate for evaluating detectors on data that reflect practical use cases of editors integrating LLMs into their editorial workflows. 
This is essential for understanding the capacity of automatic detection methods to safeguard Wikipedia’s knowledge integrity and to assist editors in identifying and removing low-quality MGT.

% The resulting potential lack of generalisability 

% hampers effective and reliable detection of MGT, 

% This highlights the need to e

% which is essential to safeguarding Wikipedia’s knowledge integrity and supporting editors in identifying and removing low-quality MGT.  

% To address this issue, we advocate for evaluating detectors on data that reflect practical use cases of how editors integrate LLMs into their editorial workflows.

To this end, we build an MGT detection benchmark for \textit{task-specific editing} scenarios on Wikipedia.  
To create our benchmark, we construct and release two new Wikipedia text corpora covering three languages with varying resource availability, enabling conclusions beyond the predominantly studied English Wikipedia.  
We then propose three editing tasks—\textit{Paragraph Writing}, \textit{Summarisation}, and \textit{Text Style Transfer}—grounded in practical use cases identified by~\citet{ford2023wiki}, who analysed Wikipedia editors' perceived opportunities for LLM-assisted editing.  
For each task, we test various prompting techniques, generate MGT using diverse LLMs, and benchmark SOTA detectors across languages, generators, and tasks (see Figure~\ref{fig:intro}). We hope that our multipurpose datasets will benefit the broader Wikipedia and AI communities in areas such as multilingual bias detection and single-document summarisation.  
We further aim to offer insights into the feasibility and reliability of automated detection methods for identifying MGT on Wikipedia.

Our contributions are as follows:
\begin{itemize}
    \item We build two datasets for our benchmark covering English, Portuguese, and Vietnamese:~\textbf{WikiPS}, a large-scale collection of high-quality (\textit{i}) lead–infobox–body triplets and (\textit{ii}) paragraphs; and~\textbf{mWNC}, an extension of the WNC~\cite{pryzant2020} to Portuguese and Vietnamese, and one of the first to include paragraph-level pairs for English.
    \item \textbf{W}ikipedia~\textbf{E}diting~\textbf{T}asks~\textbf{B}enchmark, a comprehensive benchmark of 101{,}940 \textit{task-specific} human-written and machine-generated Wikipedia texts, comprising three languages with varying levels of resource availability, four generators from two model families, and eight SOTA detectors from three detection families. We release all data and code on \href{https://github.com/gerritq/WETBenchmark}{GitHub} and plan to extend the benchmark with additional tasks, languages, and generators.
    \item We benchmark SOTA detectors on our data and find that detectors across all families struggle across tasks. While training-based detectors consistently outperform zero-shot methods, we observe substantial performance variation across languages, generators, and tasks.
\end{itemize}

% The rest of the paper is organized as follows. 
% Section~\ref{sec:relwork} discusses related work. 
% Section~\ref{sec:data} introduces our datasets. 
% Section~\ref{sec:task} describes our task definitions and prompt evaluation. Section~\ref{sec:exp} explains our experimental setup. Section~\ref{sec:res} presents the results of benchmarking detectors on our data, and section~\ref{sec:conc} concludes.

\section{Related Work}
\label{sec:relwork}

\paragraph{Wikipedia Editing Tasks} We concentrate on three common editing tasks with varying degrees of LLM involvement: Paragraph Writing, Summarisation, and Text Style Transfer.

\textit{Paragraph Writing} Generating new, encyclopaedic content---such as full paragraphs---is central to expanding knowledge on Wikipedia. This includes writing paragraphs from scratch, expanding article stubs, or rewriting existing content.  
With nearly half of all Wikipedia articles classified as stubs, researchers have extensively studied Wikipedia content generation.\footnote{\url{https://en.wikipedia.org/wiki/Wikipedia:Stub}}
The scope of generated content varies from paragraph-level~\cite[e.g.,][]{liu2018gen,balepur2023,qian2023} to full-article generation~\cite[e.g.,][]{sauper2009,banerjee2015,fan2022,shao2024,zhang2025}.  
The methods employed range from early template-based approaches~\cite{sauper2009} to more recent work using retrieval-augmented generation (RAG) with pre-trained language models (PLMs)~\cite{fan2022} or LLMs~\cite{shao2024,zhang2025}.

\textit{Summarisation} According to Wikipedia's Manual of Style,\footnote{\url{https://en.wikipedia.org/wiki/Wikipedia:Manual_of_Style}} each article should begin with a lead section that serves as an introduction by summarising its most important points.  
The literature treats lead section generation either as a multi-document~\cite[e.g.,][]{liu2018gen,ghalandari2020,hayashi2021} or single-document~\cite[e.g.,][]{casola2021,gao2021biogen,perez2022,sakota2022} summarisation problem.  
A model's objective is typically abstractive summarisation, that is, generating a lead section from scratch based on the article body.

\textit{Text Style Transfer} Maintaining a Neutral Point of View\footnote{\url{https://en.wikipedia.org/wiki/Wikipedia:Neutral_point_of_view}} (NPOV) is a core Wikipedia policy, which states that all content must be written from a perspective that is fair, proportionate, and, as far as possible, free from editorial bias.  
~\citet{pryzant2020} introduce the Wikipedia Neutrality Corpus (WNC), a large-scale parallel corpus of biased and neutralised sentence pairs retrieved from NPOV-related revisions. They further introduce the task of \textit{neutralisation}, a text style transfer task that aims to reduce subjectivity in a sentence while preserving its meaning.  
Recent work has used the WNC to improve data quality~\cite{zhong2021wikibias}, test generalisation to other domains~\cite{salas2024wikibias}, or examine the ability of LLMs to detect and neutralise bias~\cite{ashkinaze2024seeing}.

% Benchmarks (chronological order)
% TuringBench (Sept 2021): as one of the first
% Multitude (Oct 2023): first that extends AID to other languages
% M4 (March 2024): first that combines domains, langs, generators
% MAGE (May 2024):
% M4GT (August 2024): extends M4

% Check also the dataset in "A Survey on detection of..."
% There are some others: MGTBench 

% 'Standard'/'common' benchmarks
\paragraph{MGT Detection Benchmarks} There has been extensive work on benchmarking SOTA MGT detectors across diverse domains, languages, and generators. TuringBench~\cite{uchendu2021} is one of the first benchmarks to study the Turing test and authorship attribution, using multiple generators in the news domain.  
MULTITuDE~\cite{macko2023} expands MGT data for languages other than English, testing detectors in multilingual settings.  
MAGE~\cite{li2023mage} covers multiple domains, generators, and detectors, benchmarked across eight increasingly challenging detection scenarios.  
M4~\cite{wang2023m4} comprehensively includes various generators, languages, and domains, while M4GT~\cite{wang-etal-2024-m4gt} expands on M4 by incorporating additional languages and introducing human-machine mixed detection.  
A recent line of work focuses on evading detectors through adversarial attacks~\cite[e.g.,][]{he2024mgtbench,wu2024detectrl,zheng2025th}.

Most prior work has treated MGT generation primarily (\textit{i}) as an open-ended text task, (\textit{ii}) left different prompting techniques unexplored, and (\textit{iii}) produced full articles with a single prompt.  
CUDRT~\cite{tao2024cudrt} is a notable exception addressing (\textit{i}) by introducing a bilingual, multi-domain benchmark that covers five types of LLM operations.  
However, it does not consider Wikipedia, lacks analysis of how different prompting techniques affect these operations, and is limited to only three detectors.

\section{Dataset Construction}
\label{sec:data}

We construct two corpora for three languages with varying resource levels: English (high), Portuguese (medium), and Vietnamese (low).~\textbf{WikiPS} includes paragraphs and lead–content pairs.~\textbf{mWNC} is a multilingual version of the WNC~\cite{pryzant2020}. Appendix~\ref{app:cons} provides detailed descriptions of the dataset construction, and Appendix Table~\ref{tab:datastats} presents dataset statistics.

\subsection{WikiPS}  
We construct \textbf{Wiki}pedia \textbf{P}aragraphs and \textbf{S}ummarisation, a large-scale collection of Wikipedia paragraphs and lead–content pairs. To ensure that our data is not contaminated by MGT, we use the latest versions of all mainspace articles prior to the release of ChatGPT on 30 November 2022. For each language, we randomly retrieve 100{,}000 non-stub articles from the MediaWiki Action API,\footnote{\url{https://www.mediawiki.org/wiki/API:Main_page}} apply extensive filtering and cleaning of the HTML, and parse the lead section, infobox, paragraphs, and references. This forms our article-level base sample, from which we construct the paragraph and summarisation subsets, respectively.

\paragraph{Paragraphs} For each language, we consider all paragraphs from 20,000 articles in our base sample.  
To ensure paragraph quality, we retain only those that contain at least three sentences and 20 characters, include at least one reference, and have word counts within two standard deviations of the respective sample mean.  
We also add diverse metadata, such as the paragraph's location on the page, to enable filtering for specific types of paragraphs.

%Specifically define a~\textit{Introductory Paragraph}, which refers to the first paragraph following a section header. 
%\textit{Continuation Paragraph} denotes a pair consisting of the first and second halves of a paragraph, split at the sentence level, with all sentences after the median word position assigned to the second half.

\paragraph{Summarisation} We retrieve lead–infobox–body triplets from all articles in each language, as information in the lead section is often sourced from the infobox~\cite{gao2021biogen}. If an infobox is not available, we still extract the article, leaving the infobox field empty. We then merge the infobox (if present) and article body with minimal formatting into lead–content pairs.  
For English and Portuguese, we exclude pairs in which the lead/content is shorter than 10/100 characters, respectively, or longer than two standard deviations above the sample mean. For Vietnamese, we adjust the upper context limit to a minimum of 2,900 words due to its considerably longer articles.  
Appendix Table~\ref{tab:sumcomp} compares our dataset to commonly used summarisation datasets.

\subsection{mWNC} 
\textbf{m}ultilingual~\textbf{WNC} extends the original WNC~\cite{pryzant2020}, which consists of English biased–neutralised sentence pairs, by adding pairs for Portuguese and Vietnamese, as well as paragraph-level pairs for English.  
We primarily follow the methodology of~\citet{pryzant2020}, including crawling NPOV-related revisions, aligning pre- and post-neutralisation sentences, and applying rule-based filtering to improve precision.  
However, we modify their procedure by relaxing certain constraints to increase the number of instances for the Vietnamese Wikipedia, where the number of NPOV-related revisions is comparatively low.  
Furthermore, we are among the first to collect biased–neutralised paragraph-level pairs. We identify biased–neutralised paragraph pairs if three or more adjacent sentences each contain at least one NPOV-related edit.  
Due to the considerably smaller number of NPOV-related revisions in the other languages, we were only able to produce paragraph-level data for English.

\section{Editing Tasks Design}
\label{sec:task}

% Grounding (here cite Ford) + cite relevant lit
% Definition: how do we define this task
% Prompting: describe the three prmpts + embedd them in the lit + justify in terms of realism and LLM-task lit + describe eval metrics
% Expplain prompt eval results

% informed by~\citet{ford2023wiki} and 
We define three editing tasks with varying degrees of LLM intervention: \textit{Paragraph Writing}, \textit{Summarisation}, and \textit{Text Style Transfer}.  
These tasks are empirically motivated by~\citet{ford2023wiki}, who found that Wikipedia editors see potential in LLMs for \textit{generating article drafts or stubs}, \textit{summarising content}, and \textit{improving language}.  
We implement Paragraph Writing and Summarisation using the WikiPS corpus, and Text Style Transfer using the mWNC.

For each task and language, we evaluate three prompting strategies on a length-stratified 10\% sample of the target data using GPT-4o mini,\footnote{\url{https://platform.openai.com/docs/models/gpt-4o-mini}} and select the best-performing prompt to generate MGT for our benchmark. Appendix~\ref{app:tasks} provides implementation details and prompt templates.

\subsection{Paragraph Writing}
\label{sec:pwriting}

We define~\textit{Paragraph Writing} as the task of writing the opening paragraph of a new section, resembling a scenario in which an editor aims to add new content to an article. In contrast to prior work on open-ended generation~\cite[e.g.,][]{guo2023closechatgpthumanexperts,li2023mage,wang2023m4,wang2024m4gt}, we frame this as a \textit{content-conditioned} generation task, where the model receives additional information about the content and style of the output.  
This \textit{content creation} task involves the highest degree of LLM contribution, as the model generates the paragraph from scratch.

We devise three prompts with increasing levels of content conditioning.~\textbf{Minimal} simply instructs the model to write a paragraph given article and section titles. We include this prompt as it reflects generation settings in prior work and thus serves as a comparative baseline.~\textbf{Content Prompts} expand Minimal by incorporating up to ten content prompts about the target HWT paragraph (e.g., "What is London's population?"), obtained from GPT-4o,\footnote{\url{https://openai.com/index/gpt-4o-system-card/}} to steer the model towards factual alignment with the HWT reference.  
Lastly, to enhance the factual accuracy of the generated text, we implement a web-based search \textbf{Naive RAG}~\cite{gao2024rag}, which adds relevant context to the Content Prompts. Appendix~\ref{app:rag} provides implementation details of Naive RAG.

We evaluate these prompts using standard automatic metrics: BLEU~\cite{papineni2002bleu} and ROUGE~\cite{lin2004rouge} for n-gram overlap, BERTScore~\cite{zhang2019bertscore} for semantic similarity, and QAFactEval~\cite{fabbri2022} (F1-score) as a QA-based metric for factual consistency between HWT and MGT.\footnote{For Portuguese and Vietnamese texts, QAFactEval evaluations were performed using GPT-4 translations.}

\begin{table}[!htbp]
    \centering
    \large
    \begin{adjustbox}{width=\linewidth}
\begin{tabular}{llcccccc}
    \toprule
     \textbf{Language} & \textbf{Technique} & \textbf{BLEU} & \textbf{ROUGE-1} & \textbf{ROUGE-2} & \textbf{ROUGE-L} & \textbf{BERTScore} & \textbf{QAFactEval} \\
    \midrule
    % \midrule
    % \multicolumn{8}{l}{\textit{Introductory Paragraph}} \\
    % \midrule
    \multirow{3}{*}{English} & Minimal & 0.02 & 0.29 & 0.06 & 0.17 & 0.76 & 0.06 \\
     & Content Prompts & 0.22 & 0.57 & 0.31 & 0.44 & \textbf{0.88} & 0.25 \\
     & RAG & \textbf{0.2}5 & \textbf{0.61} & \textbf{0.35} & \textbf{0.47} & \textbf{0.88} & \textbf{0.38} \\
    \cmidrule{1-8}
    \multirow{3}{*}{Portuguese} & Minimal & 0.02 & 0.31 & 0.06 & 0.17 & 0.86 & 0.06 \\
     & Content Prompts & 0.20 & 0.56 & 0.30 & 0.41 & 0.91 & 0.25 \\
     & RAG & \textbf{0.25} & \textbf{0.61} & \textbf{0.37} & \textbf{0.47} & \textbf{0.92} & \textbf{0.42} \\
    \cmidrule{1-8}
    \multirow{3}{*}{Vietnamese} & Minimal & 0.04 & 0.67 & 0.26 & 0.32 & 0.85 & 0.06 \\
     & Content Prompts & 0.28 & 0.78 & 0.52 & 0.54 & 0.91 & 0.27 \\
     & RAG & \textbf{0.30} & \textbf{0.79} & \textbf{0.54} & \textbf{0.55} & \textbf{0.92} & \textbf{0.36} \\
    % \midrule
    % \midrule
    % \multicolumn{8}{l}{\textit{Paragraph Continuation}} \\
    % \midrule
    % \multirow{3}{*}{English} & Minimal & 0.01 & 0.24 & 0.03 & 0.15 & 0.75 & 0.03 \\
    %  & Content Prompts & 0.21 & 0.58 & 0.32 & 0.45 & 0.88 & 0.30 \\
    %  & RAG & \textbf{0.25} & \textbf{0.60} & \textbf{0.36} & \textbf{0.49} & 0.89 & \textbf{0.42} \\
    % \cmidrule{1-8}
    % \multirow{3}{*}{Portuguese} & Minimal & 0.01 & 0.25 & 0.04 & 0.15 & 0.86 & 0.03 \\
    %  & Content Prompts & 0.20 & 0.57 & 0.32 & 0.44 & 0.92 & 0.27 \\
    %  & RAG & \textbf{0.25} & \textbf{0.60} & \textbf{0.38} & \textbf{0.49} & 0.92 & \textbf{0.42} \\
    % \cmidrule{1-8}
    % \multirow{3}{*}{Vietnamese} & Minimal & 0.01 & 0.62 & 0.21 & 0.31 & 0.85 & 0.04 \\
    %  & Content Prompts & 0.31 & 0.78 & 0.54 & 0.56 & 0.92 & 0.31 \\
    %  & RAG & 0.32 & 0.78 & 0.54 & 0.57 & 0.92 & \textbf{0.38} \\
    % \bottomrule
    \bottomrule
\end{tabular}
    \end{adjustbox}
    
    \caption{Paragraph Writing prompts evaluation results.}
    \label{tab:pprompts}

\end{table}

Table~\ref{tab:pprompts} presents our prompting evaluation results. We find that our Naive RAG approach consistently outperforms both Minimal and Content Prompts across subtasks and languages. The low evaluation scores for Minimal prompts highlight that MGT produced in prior work is often synthetically divergent from its human-written references. While Content Prompts substantially improve performance, Naive RAG further enhances generation quality, particularly in terms of factual consistency, which is critical for encyclopaedic content.\footnote{\url{https://en.wikipedia.org/wiki/Wikipedia:Verifiability}}  
Based on these findings, we adopt Naive RAG as the prompting strategy for the Paragraph Writing task in our MGT detection experiments.

\subsection{Summarisation}

\textit{Summarisation} tasks the model with generating a lead section of comparable length to the human-written reference, based on the article's body and infobox, both of which are the main sources for lead section information~\cite{gao2021biogen}.  
We frame this as a single-document, abstractive summarisation task, following Wikipedia’s Manual of Style\footnote{\url{https://en.wikipedia.org/wiki/Wikipedia:Manual_of_Style/Lead_section}} and prior work on Wikipedia summarisation~\cite{casola2021,gao2021biogen,perez2022}.  
Compared to \textit{Paragraph Writing}, this \textit{content condensation} task involves slightly less LLM contribution due to its stronger grounding in existing article content.

We use three prompting techniques from the literature on LLM-generated summaries~\cite{goyal2022news,pu2023summarization,zhang2023bench} that align with this editing scenario. Each prompt contains the article content as input and conditions the output length on the target lead length.  
\textbf{Minimal} is a simple zero-shot baseline prompt that instructs the model to summarise the article content. \textbf{Instruction} adds a concise definition of, and instructions for compiling, a lead section to guide the model more explicitly. \textbf{Few-shot} further includes 1–3 high-quality lead–content examples, retrieved from the respective Wikipedia Featured Articles page, in addition to the Instruction prompt to enable in-context learning~\cite{brown2020}.\footnote{\url{https://en.wikipedia.org/wiki/Wikipedia:Featured_articles}}  
We evaluate these prompts using traditional automatic metrics for summarisation evaluation (see Section~\ref{sec:pwriting}).

\begin{table}[h]
    \centering
    \footnotesize
    \begin{adjustbox}{width=\linewidth}

    % INSERT TABLE HERE
    \begin{tabular}{llrrrrrr}
    \toprule
    \textbf{Language} & \textbf{Technique} & \textbf{BLEU} & \textbf{ROUGE-1} & \textbf{ROUGE-2} & \textbf{ROUGE-L} & \textbf{BERTScore} & \textbf{QAFactEval} \\
    \midrule
    English & Minimal & 0.06 & 0.37 & 0.13 & 0.26 & 0.79 & 0.45 \\
     & Instruction & 0.13 & 0.44 & 0.21 & 0.33 & 0.82 & \textbf{0.46} \\
     & One-shot & \textbf{0.18} & \textbf{0.47} & \textbf{0.24} & \textbf{0.36} & \textbf{0.83} & \textbf{0.46} \\
     & Two-shot & \textbf{0.18} & \textbf{0.47} & \textbf{0.24} & \textbf{0.36} & \textbf{0.83} & \textbf{0.46} \\
     & Three-shot & 0.16 & 0.46 & 0.23 & 0.35 & \textbf{0.83} & \textbf{0.46} \\
    \midrule
    Portuguese & Minimal & 0.06 & 0.35 & 0.13 & 0.23 & 0.87 & \textbf{0.48} \\
     & Instruction & \textbf{0.11} & 0.42 & 0.19 & 0.30 & \textbf{0.88} & \textbf{0.48} \\
     & One-shot & \textbf{0.11} & 0.42 & 0.19 & 0.29 & \textbf{0.88} & \textbf{0.48} \\
     & Two-shot & \textbf{0.11} & \textbf{0.43} & 0.19 & \textbf{0.30} & \textbf{0.88} & 0.47 \\
     & Three-shot & 0.12 & \textbf{0.43} & \textbf{0.20} & \textbf{0.30} & \textbf{0.88} & 0.47 \\
    \midrule
    Vietnamese & Minimal & 0.07 & 0.63 & 0.28 & 0.35 & 0.86 &\textbf{ 0.45} \\
     & Instruction & 0.11 & 0.64 & 0.31 & \textbf{0.38} & \textbf{0.87} & 0.43 \\
     & One-shot & \textbf{0.12} & 0.65 & \textbf{0.32} & \textbf{0.38} & \textbf{0.87} & \textbf{0.45} \\
     & Two-shot & \textbf{0.12} & \textbf{0.66} & \textbf{0.32} & \textbf{0.38} & \textbf{0.87} & 0.44 \\
     & Three-shot & 0.11 & 0.65 & \textbf{0.32} & \textbf{0.38} & \textbf{0.87} & 0.42 \\
    \midrule
    \end{tabular}

    \end{adjustbox}
    \caption{Summarisation prompts evaluation results.}
    \label{tab:sprompts}
\end{table}

Table~\ref{tab:sprompts} presents the summarisation prompt evaluation results, showing that across languages, Instruction and Few-shot achieve higher overlap and semantic similarity scores, although Few-shot only marginally improves over Instruction.  
Factuality scores remain relatively stable across prompts, presumably because summarisation is a core task in aligning LLMs through reinforcement learning from human feedback~\cite{ouyang2022}.  
Given that increasing the number of shots does not yield further improvements, and considering the context window of smaller LLMs, we select one-shot prompting for our experiments.

\subsection{Text Style Transfer}
\label{sec:tst}

We adopt the TST task of \textit{neutralising} revision-level NPOV violations, as introduced by~\citet{pryzant2020}. In our setup, the model is instructed to revise a biased sentence or paragraph with minimal edits, aligning the output with Wikipedia’s neutrality guidelines.  
While various TST tasks are possible on Wikipedia, focusing on NPOV violations ensures direct alignment with one of its core content policies.\footnote{\url{https://en.wikipedia.org/wiki/Wikipedia:Neutral_point_of_view}}  
This \textit{content modification} task involves the least LLM contribution, as the model is conditioned to perform only minor revisions to existing text.

We test three prompting techniques for TST that are conceptually identical to those used in summarisation and align with recent work on LLM-based TST~\cite{reif2021recipe,dwivedi2022editeval,ashkinaze2024seeing}.  
All prompts include the biased input text and constrain the output to be no longer than the target text.  
Compared to summarisation, \textbf{Minimal} instructs the model to neutralise the input; \textbf{Instruction} adds a concise definition of Wikipedia’s NPOV policy; and \textbf{Few-shot} includes 1–5 randomly sampled biased–neutralised examples.

We evaluate these TST prompts along two dimensions: \textit{semantic content preservation}, for which we report BLEU~\cite{papineni2002bleu}, ROUGE~\cite{lin2004rouge}, and BERTScore~\cite{zhang2019bertscore}; and \textit{style transfer accuracy}, for which we fine-tune pre-trained language models for each language and report the accuracy of binary style classification.  
Fine-tuning details are provided in Appendix~\ref{app:tst}.

\begin{table}[h]
    \centering
    \begin{adjustbox}{width=\linewidth}

    % INSERT TABLE HERE
    \begin{tabular}{llcccccc}
    \toprule
    \textbf{Language} & \textbf{Technique} & \textbf{BLEU} & \textbf{ROUGE-1} & \textbf{ROUGE-2} & \textbf{ROUGE-L} & \textbf{BERTScore} & \textbf{ST} \\
    \midrule
    English & Minimal & 0.35 & 0.68 & 0.52 & 0.66 & 0.92 & 0.90 \\
     & Instruction & 0.36 & 0.68 & 0.52 & 0.66 & 0.92 & \textbf{0.94} \\
    \cdashline{2-8}[2pt/2pt]
     & One-shot & 0.52 & 0.78 & 0.65 & 0.76 & \textbf{0.95} & 0.91 \\
     & Two-shot & 0.47 & 0.75 & 0.61 & 0.73 & 0.94 & 0.90 \\
     & Three-shot & 0.54 & 0.79 & 0.67 & 0.78 & \textbf{0.95} & 0.89 \\
     & Four-shot & \textbf{0.56} & \textbf{0.80} & \textbf{0.69} & \textbf{0.79} & \textbf{0.95} & 0.89 \\
     & Five-shot & \textbf{0.55} & \textbf{0.80} & 0.68 & 0.78 & \textbf{0.95} & 0.91 \\
    \midrule
    Portuguese & Minimal & 0.41 & 0.71 & 0.58 & 0.69 & 0.94 & 0.86 \\
     & Instruction & 0.40 & 0.70 & 0.57 & 0.67 & 0.94 & 0.88 \\
    \cdashline{2-8}[2pt/2pt]
     & One-shot & 0.50 & 0.75 & 0.64 & 0.74 & \textbf{0.96} & 0.90 \\
     & Two-shot & 0.51 & 0.77 & 0.65 & 0.75 & \textbf{0.96} & 0.89 \\
     & Three-shot & 0.53 & 0.78 & 0.66 & 0.76 & \textbf{0.96} & 0.91 \\
     & Four-shot & \textbf{0.58} & \textbf{0.81} & \textbf{0.70} & \textbf{0.79} & \textbf{0.96} & \textbf{0.92} \\
     & Five-shot & 0.55 & 0.79 & 0.68 & 0.77 & \textbf{0.96} & 0.91 \\
    \midrule
    Vietnamese & Minimal & 0.43 & 0.78 & 0.65 & 0.73 & 0.95 & 0.84 \\
     & Instruction & 0.45 & 0.80 & 0.67 & 0.73 & 0.94 & 0.79 \\
    \cdashline{2-8}[2pt/2pt]
     & One-shot & 0.44 & 0.78 & 0.66 & 0.71 & 0.95 & \textbf{0.88} \\
     & Two-shot & 0.51 & 0.82 & 0.70 & 0.76 & 0.95 & 0.87 \\
     & Three-shot & 0.50 & 0.81 & 0.70 & 0.75 & 0.95 & 0.85 \\
     & Four-shot & 0.51 & 0.82 & 0.70 & 0.76 & 0.95 & 0.85 \\
     & Five-shot & \textbf{0.55} & \textbf{0.83} & \textbf{0.73} & \textbf{0.78} & \textbf{0.96} & 0.84 \\
    \midrule
    \makecell{English Para.} & Minimal & 0.35 & 0.68 & 0.52 & 0.66 & 0.92 & 0.97 \\
     & Instruction & 0.36 & 0.68 & 0.52 & 0.66 & 0.92 & \textbf{0.99} \\
    \cdashline{2-8}[2pt/2pt]
     & One-shot & 0.52 & 0.78 & 0.65 & 0.76 &\textbf{ 0.95} & 0.95 \\
     & Two-shot & 0.47 & 0.75 & 0.61 & 0.73 & 0.94 & 0.98 \\
     & Three-shot & 0.54 & 0.79 & 0.67 & 0.78 & \textbf{0.95} & 0.96 \\
     & Four-shot & \textbf{0.56} & \textbf{0.80} & \textbf{0.69} & \textbf{0.79} & \textbf{0.95} & 0.95 \\
     & Five-shot & 0.55 & \textbf{0.80} & 0.68 & 0.78 & \textbf{0.95} & 0.96 \\
    \bottomrule
    \end{tabular}    
    \end{adjustbox}
    \caption{TST prompts evaluation results.}
    \label{tab:tstprompts}
\end{table}

Table~\ref{tab:tstprompts} presents the prompt evaluation metrics for the TST task, evaluated at the sentence level for all languages, and additionally at the paragraph level for English.  
Across languages and levels, we find that four- and five-shot prompting consistently outperforms Minimal and Instruction prompts.  
While differences in semantic similarity and style transfer are marginal across prompts, we observe substantial improvements in overlap-based metrics as the number of few-shot examples increases.  
These improvements can be attributed to the fact that neutralisation edits in mWNC tend to be relatively minimal. For instance, in the English sentence subset, on average only 14\% of words are deleted and 7\% added—similar trends hold for the other subsets.  
As a result, the model appears to learn from the examples to apply similarly sparse edits, thereby producing outputs that match the reference text more closely in terms of n-gram overlap.  
Based on these findings, we adopt five-shot prompting to generate MGT in our subsequent experiments.

\begin{table*}[!htbp]
    \centering
    \begin{adjustbox}{width=\textwidth}

\begin{tabular}{llcccccccccc p{.1in} cccccccccc p{.1in} cccccccccc}
\toprule
\textbf{Task} & \textbf{Detector} & \multicolumn{10}{c}{\textbf{English}} &  & \multicolumn{10}{c}{\textbf{Portuguese}} &  & \multicolumn{10}{c}{\textbf{Vietnamese}} \\
\cmidrule(lr){3-12} \cmidrule(lr){14-23} \cmidrule(lr){25-34}
 &  & \multicolumn{2}{c}{GPT-4o mini} & \multicolumn{2}{c}{Gemini 2.0} & \multicolumn{2}{c}{Qwen 2.5} & \multicolumn{2}{c}{Mistral} & \multicolumn{2}{c}{\textbf{Avg}} &  & \multicolumn{2}{c}{GPT-4o mini} & \multicolumn{2}{c}{Gemini 2.0} & \multicolumn{2}{c}{Qwen 2.5} & \multicolumn{2}{c}{Mistral} & \multicolumn{2}{c}{\textbf{Avg}} &  & \multicolumn{2}{c}{GPT-4o mini} & \multicolumn{2}{c}{Gemini 2.0} & \multicolumn{2}{c}{Qwen 2.5} & \multicolumn{2}{c}{Mistral} & \multicolumn{2}{c}{\textbf{Avg}} \\
\cmidrule(lr){3-4} \cmidrule(lr){5-6} \cmidrule(lr){7-8} \cmidrule(lr){9-10} \cmidrule(lr){11-12} \cmidrule(lr){14-15} \cmidrule(lr){16-17} \cmidrule(lr){18-19} \cmidrule(lr){20-21} \cmidrule(lr){22-23} \cmidrule(lr){25-26} \cmidrule(lr){27-28} \cmidrule(lr){29-30} \cmidrule(lr){31-32} \cmidrule(lr){33-34}
 &  & ACC & F1 & ACC & F1 & ACC & F1 & ACC & F1 & ACC & F1 &  & ACC & F1 & ACC & F1 & ACC & F1 & ACC & F1 & ACC & F1 &  & ACC & F1 & ACC & F1 & ACC & F1 & ACC & F1 & ACC & F1 \\
\midrule
\multirow{11}{*}{Introductory Paragraph} & Binoculars & 0.61 & 0.60 & 0.58 & 0.61 & 0.60 & 0.58 & 0.55 & 0.63 & \textbf{\greygra{0.59}} & \textbf{\greygra{0.60}} &  & 0.68 & 0.66 & 0.64 & 0.64 & 0.64 & 0.60 & 0.54 & 0.61 & \textbf{\greygra{0.62}} & \textbf{\greygra{0.63}} &  & 0.77 & 0.77 & 0.72 & 0.72 & 0.70 & 0.66 & 0.50 & 0.67 & \textbf{\greygra{0.67}} & \textbf{\greygra{0.70}} \\
 & LLR & 0.52 & 0.51 & 0.50 & 0.67 & 0.50 & 0.67 & 0.53 & 0.62 & \textbf{\greygra{0.51}} & \textbf{\greygra{0.62}} &  & 0.57 & 0.51 & 0.54 & 0.51 & 0.50 & 0.67 & 0.53 & 0.51 & \textbf{\greygra{0.53}} & \textbf{\greygra{0.55}} &  & 0.63 & 0.60 & 0.58 & 0.54 & 0.51 & 0.19 & 0.50 & 0.00 & \textbf{\greygra{0.56}} & \textbf{\greygra{0.33}} \\
 & FDGPT (WB) & 0.59 & 0.60 & 0.54 & 0.52 & 0.52 & 0.43 & 0.52 & 0.52 & \textbf{\greygra{0.54}} & \textbf{\greygra{0.51}} &  & 0.68 & 0.66 & 0.63 & 0.63 & 0.56 & 0.53 & 0.52 & 0.57 & \textbf{\greygra{0.60}} & \textbf{\greygra{0.60}} &  & 0.76 & 0.77 & 0.70 & 0.69 & 0.59 & 0.53 & 0.50 & 0.00 & \textbf{\greygra{0.64}} & \textbf{\greygra{0.50}} \\
\cdashline{2-34} \addlinespace[1pt]
 & Avg. White-box & \greygra{0.57} & \greygra{0.57} & \greygra{0.54} & \greygra{0.60} & \greygra{0.54} & \greygra{0.56} & \greygra{0.54} & \greygra{0.59} & \textbf{\greygra{0.55}} & \textbf{\greygra{0.58}} &  & \greygra{0.64} & \greygra{0.61} & \greygra{0.60} & \greygra{0.59} & \greygra{0.56} & \greygra{0.60} & \greygra{0.53} & \greygra{0.56} & \textbf{\greygra{0.58}} & \textbf{\greygra{0.59}} &  & \greygra{0.72} & \greygra{0.71} & \greygra{0.67} & \greygra{0.65} & \greygra{0.60} & \greygra{0.46} & \greygra{0.50} & \greygra{0.22} & \textbf{\greygra{0.62}} & \textbf{\greygra{0.51}} \\
\addlinespace[3pt]
 & Revise & 0.53 & 0.41 & 0.53 & 0.50 & 0.52 & 0.55 & 0.52 & 0.62 & \textbf{\greygra{0.52}} & \textbf{\greygra{0.52}} &  & 0.55 & 0.58 & 0.56 & 0.50 & 0.54 & 0.53 & 0.52 & 0.59 & \textbf{\greygra{0.54}} & \textbf{\greygra{0.55}} &  & 0.53 & 0.50 & 0.54 & 0.56 & 0.54 & 0.59 & 0.50 & 0.00 & \textbf{\greygra{0.53}} & \textbf{\greygra{0.41}} \\
 & GECScore & 0.82 & 0.82 & 0.77 & 0.75 & 0.77 & 0.75 & 0.72 & 0.73 & \textbf{\greygra{0.77}} & \textbf{\greygra{0.76}} &  & 0.79 & 0.79 & 0.80 & 0.80 & 0.65 & 0.66 & 0.54 & 0.66 & \textbf{\greygra{0.69}} & \textbf{\greygra{0.73}} &  & 0.72 & 0.70 & 0.70 & 0.70 & 0.58 & 0.47 & 0.50 & 0.67 & \textbf{\greygra{0.62}} & \textbf{\greygra{0.63}} \\
 & FDGPT (BB) & 0.58 & 0.61 & 0.53 & 0.49 & 0.52 & 0.45 & 0.54 & 0.50 & \textbf{\greygra{0.55}} & \textbf{\greygra{0.51}} &  & 0.69 & 0.66 & 0.62 & 0.60 & 0.56 & 0.44 & 0.54 & 0.42 & \textbf{\greygra{0.60}} & \textbf{\greygra{0.53}} &  & 0.74 & 0.74 & 0.68 & 0.70 & 0.59 & 0.59 & 0.50 & 0.00 & \textbf{\greygra{0.63}} & \textbf{\greygra{0.51}} \\
\cdashline{2-34} \addlinespace[1pt]
 & Avg. Black-box & \greygra{0.64} & \greygra{0.61} & \greygra{0.61} & \greygra{0.58} & \greygra{0.60} & \greygra{0.58} & \greygra{0.59} & \greygra{0.62} & \textbf{\greygra{0.61}} & \textbf{\greygra{0.60}} &  & \greygra{0.68} & \greygra{0.68} & \greygra{0.66} & \greygra{0.63} & \greygra{0.58} & \greygra{0.54} & \greygra{0.53} & \greygra{0.56} & \textbf{\greygra{0.61}} & \textbf{\greygra{0.60}} &  & \greygra{0.66} & \greygra{0.65} & \greygra{0.64} & \greygra{0.65} & \greygra{0.57} & \greygra{0.55} & \greygra{0.50} & \greygra{0.22} & \textbf{\greygra{0.59}} & \textbf{\greygra{0.52}} \\
\addlinespace[3pt]
 & xlm-RoBERTa & 0.81 & 0.81 & 0.84 & 0.84 & 0.85 & 0.85 & 0.83 & 0.83 & \textbf{\greygra{0.83}} & \textbf{\greygra{0.83}} &  & 0.76 & 0.75 & 0.80 & 0.79 & 0.83 & 0.82 & 0.76 & 0.74 & \textbf{\greygra{0.79}} & \textbf{\greygra{0.78}} &  & 0.75 & 0.72 & 0.82 & 0.81 & 0.89 & 0.89 & 0.98 & 0.98 & \textbf{\greygra{0.86}} & \textbf{\greygra{0.85}} \\
 & mDeBERTa & 0.86 & 0.86 & 0.84 & 0.83 & 0.89 & 0.88 & 0.90 & 0.89 & \textbf{\greygra{0.87}} & \textbf{\greygra{0.87}} &  & 0.76 & 0.74 & 0.80 & 0.79 & 0.84 & 0.83 & 0.85 & 0.85 & \textbf{\greygra{0.81}} & \textbf{\greygra{0.80}} &  & 0.77 & 0.76 & 0.82 & 0.81 & 0.87 & 0.87 & 0.98 & 0.98 & \textbf{\greygra{0.86}} & \textbf{\greygra{0.86}} \\
\cdashline{2-34} \addlinespace[1pt]
 & Avg. Supervised & \greygra{0.84} & \greygra{0.83} & \greygra{0.84} & \greygra{0.84} & \greygra{0.87} & \greygra{0.87} & \greygra{0.86} & \greygra{0.86} & \textbf{\greygra{0.85}} & \textbf{\greygra{0.85}} &  & \greygra{0.76} & \greygra{0.75} & \greygra{0.80} & \greygra{0.79} & \greygra{0.83} & \greygra{0.83} & \greygra{0.80} & \greygra{0.79} & \textbf{\greygra{0.80}} & \textbf{\greygra{0.79}} &  & \greygra{0.76} & \greygra{0.74} & \greygra{0.82} & \greygra{0.81} & \greygra{0.88} & \greygra{0.88} & \greygra{0.98} & \greygra{0.98} & \textbf{\greygra{0.86}} & \textbf{\greygra{0.85}} \\
\addlinespace[3pt]
\midrule
\multirow{11}{*}{Summarisation} & Binoculars & 0.60 & 0.60 & 0.62 & 0.58 & 0.62 & 0.62 & 0.62 & 0.69 & \textbf{\greygra{0.61}} & \textbf{\greygra{0.62}} &  & 0.72 & 0.73 & 0.70 & 0.72 & 0.71 & 0.72 & 0.62 & 0.66 & \textbf{\greygra{0.69}} & \textbf{\greygra{0.71}} &  & 0.72 & 0.72 & 0.70 & 0.71 & 0.72 & 0.73 & 0.50 & 0.67 & \textbf{\greygra{0.66}} & \textbf{\greygra{0.71}} \\
 & LLR & 0.52 & 0.65 & 0.52 & 0.64 & 0.54 & 0.66 & 0.61 & 0.68 & \textbf{\greygra{0.55}} & \textbf{\greygra{0.66}} &  & 0.58 & 0.57 & 0.56 & 0.54 & 0.54 & 0.65 & 0.55 & 0.67 & \textbf{\greygra{0.56}} & \textbf{\greygra{0.61}} &  & 0.58 & 0.47 & 0.55 & 0.47 & 0.54 & 0.65 & 0.51 & 0.67 & \textbf{\greygra{0.55}} & \textbf{\greygra{0.56}} \\
 & FDGPT (WB) & 0.60 & 0.59 & 0.60 & 0.59 & 0.55 & 0.51 & 0.61 & 0.60 & \textbf{\greygra{0.59}} & \textbf{\greygra{0.57}} &  & 0.72 & 0.70 & 0.69 & 0.69 & 0.65 & 0.63 & 0.56 & 0.58 & \textbf{\greygra{0.65}} & \textbf{\greygra{0.65}} &  & 0.73 & 0.72 & 0.71 & 0.71 & 0.64 & 0.62 & 0.50 & 0.00 & \textbf{\greygra{0.64}} & \textbf{\greygra{0.51}} \\
\cdashline{2-34} \addlinespace[1pt]
 & Avg. White-box & \greygra{0.57} & \greygra{0.61} & \greygra{0.58} & \greygra{0.61} & \greygra{0.57} & \greygra{0.60} & \greygra{0.61} & \greygra{0.66} & \textbf{\greygra{0.58}} & \textbf{\greygra{0.62}} &  & \greygra{0.67} & \greygra{0.67} & \greygra{0.65} & \greygra{0.65} & \greygra{0.63} & \greygra{0.67} & \greygra{0.58} & \greygra{0.64} & \textbf{\greygra{0.63}} & \textbf{\greygra{0.66}} &  & \greygra{0.68} & \greygra{0.64} & \greygra{0.66} & \greygra{0.63} & \greygra{0.64} & \greygra{0.67} & \greygra{0.50} & \greygra{0.45} & \textbf{\greygra{0.62}} & \textbf{\greygra{0.59}} \\
\addlinespace[3pt]
 & Revise & 0.53 & 0.61 & 0.53 & 0.58 & 0.53 & 0.62 & 0.53 & 0.61 & \textbf{\greygra{0.53}} & \textbf{\greygra{0.60}} &  & 0.54 & 0.51 & 0.53 & 0.57 & 0.53 & 0.55 & 0.51 & 0.63 & \textbf{\greygra{0.53}} & \textbf{\greygra{0.56}} &  & 0.53 & 0.57 & 0.54 & 0.56 & 0.54 & 0.56 & 0.50 & 0.66 & \textbf{\greygra{0.53}} & \textbf{\greygra{0.59}} \\
 & GECScore & 0.80 & 0.81 & 0.71 & 0.70 & 0.75 & 0.75 & 0.74 & 0.76 & \textbf{\greygra{0.75}} & \textbf{\greygra{0.75}} &  & 0.78 & 0.79 & 0.73 & 0.72 & 0.68 & 0.70 & 0.52 & 0.64 & \textbf{\greygra{0.68}} & \textbf{\greygra{0.71}} &  & 0.71 & 0.70 & 0.67 & 0.66 & 0.63 & 0.64 & 0.50 & 0.67 & \textbf{\greygra{0.63}} & \textbf{\greygra{0.67}} \\
 & FDGPT (BB) & 0.59 & 0.58 & 0.59 & 0.63 & 0.56 & 0.60 & 0.63 & 0.62 & \textbf{\greygra{0.59}} & \textbf{\greygra{0.61}} &  & 0.71 & 0.69 & 0.67 & 0.70 & 0.65 & 0.66 & 0.57 & 0.57 & \textbf{\greygra{0.65}} & \textbf{\greygra{0.65}} &  & 0.70 & 0.68 & 0.68 & 0.66 & 0.64 & 0.61 & 0.50 & 0.01 & \textbf{\greygra{0.63}} & \textbf{\greygra{0.49}} \\
\cdashline{2-34} \addlinespace[1pt]
 & Avg. Black-box & \greygra{0.64} & \greygra{0.66} & \greygra{0.61} & \greygra{0.63} & \greygra{0.61} & \greygra{0.66} & \greygra{0.63} & \greygra{0.66} & \textbf{\greygra{0.62}} & \textbf{\greygra{0.65}} &  & \greygra{0.67} & \greygra{0.66} & \greygra{0.64} & \greygra{0.66} & \greygra{0.62} & \greygra{0.63} & \greygra{0.53} & \greygra{0.61} & \textbf{\greygra{0.62}} & \textbf{\greygra{0.64}} &  & \greygra{0.65} & \greygra{0.65} & \greygra{0.63} & \greygra{0.63} & \greygra{0.60} & \greygra{0.60} & \greygra{0.50} & \greygra{0.45} & \textbf{\greygra{0.60}} & \textbf{\greygra{0.58}} \\
\addlinespace[3pt]
 & xlm-RoBERTa & 0.92 & 0.92 & 0.86 & 0.85 & 0.95 & 0.95 & 0.90 & 0.90 & \textbf{\greygra{0.91}} & \textbf{\greygra{0.90}} &  & 0.91 & 0.91 & 0.84 & 0.84 & 0.91 & 0.91 & 0.94 & 0.94 & \textbf{\greygra{0.90}} & \textbf{\greygra{0.90}} &  & 0.86 & 0.86 & 0.78 & 0.77 & 0.94 & 0.93 & 0.94 & 0.94 & \textbf{\greygra{0.88}} & \textbf{\greygra{0.87}} \\
 & mDeBERTa & 0.91 & 0.91 & 0.83 & 0.83 & 0.93 & 0.93 & 0.91 & 0.90 & \textbf{\greygra{0.89}} & \textbf{\greygra{0.89}} &  & 0.89 & 0.89 & 0.86 & 0.85 & 0.94 & 0.93 & 0.94 & 0.94 & \textbf{\greygra{0.91}} & \textbf{\greygra{0.90}} &  & 0.84 & 0.84 & 0.77 & 0.76 & 0.90 & 0.90 & 0.96 & 0.96 & \textbf{\greygra{0.87}} & \textbf{\greygra{0.87}} \\
\cdashline{2-34} \addlinespace[1pt]
 & Avg. Supervised & \greygra{0.91} & \greygra{0.91} & \greygra{0.84} & \greygra{0.84} & \greygra{0.94} & \greygra{0.94} & \greygra{0.90} & \greygra{0.90} & \textbf{\greygra{0.90}} & \textbf{\greygra{0.90}} &  & \greygra{0.90} & \greygra{0.90} & \greygra{0.85} & \greygra{0.85} & \greygra{0.92} & \greygra{0.92} & \greygra{0.94} & \greygra{0.94} & \textbf{\greygra{0.90}} & \textbf{\greygra{0.90}} &  & \greygra{0.85} & \greygra{0.85} & \greygra{0.78} & \greygra{0.76} & \greygra{0.92} & \greygra{0.92} & \greygra{0.95} & \greygra{0.95} & \textbf{\greygra{0.87}} & \textbf{\greygra{0.87}} \\
\addlinespace[3pt]
\midrule
\multirow{11}{*}{Text Style Transfer} & Binoculars & 0.53 & 0.46 & 0.52 & 0.33 & 0.50 & 0.05 & 0.53 & 0.54 & \textbf{\greygra{0.52}} & \textbf{\greygra{0.34}} &  & 0.56 & 0.54 & 0.55 & 0.51 & 0.53 & 0.49 & 0.52 & 0.48 & \textbf{\greygra{0.54}} & \textbf{\greygra{0.51}} &  & 0.59 & 0.55 & 0.57 & 0.52 & 0.53 & 0.47 & 0.50 & 0.00 & \textbf{\greygra{0.55}} & \textbf{\greygra{0.39}} \\
 & LLR & 0.50 & 0.02 & 0.50 & 0.05 & 0.50 & 0.02 & 0.50 & 0.04 & \textbf{\greygra{0.50}} & \textbf{\greygra{0.03}} &  & 0.51 & 0.25 & 0.50 & 0.03 & 0.50 & 0.02 & 0.50 & 0.32 & \textbf{\greygra{0.50}} & \textbf{\greygra{0.15}} &  & 0.52 & 0.29 & 0.51 & 0.16 & 0.51 & 0.26 & 0.50 & 0.67 & \textbf{\greygra{0.51}} & \textbf{\greygra{0.34}} \\
 & FDGPT (WB) & 0.54 & 0.55 & 0.53 & 0.54 & 0.50 & 0.67 & 0.53 & 0.54 & \textbf{\greygra{0.52}} & \textbf{\greygra{0.57}} &  & 0.56 & 0.53 & 0.54 & 0.51 & 0.51 & 0.44 & 0.51 & 0.49 & \textbf{\greygra{0.53}} & \textbf{\greygra{0.49}} &  & 0.58 & 0.57 & 0.55 & 0.55 & 0.53 & 0.52 & 0.50 & 0.00 & \textbf{\greygra{0.54}} & \textbf{\greygra{0.41}} \\
\cdashline{2-34} \addlinespace[1pt]
 & Avg. White-box & \greygra{0.52} & \greygra{0.34} & \greygra{0.52} & \greygra{0.31} & \greygra{0.50} & \greygra{0.24} & \greygra{0.52} & \greygra{0.37} & \textbf{\greygra{0.52}} & \textbf{\greygra{0.32}} &  & \greygra{0.54} & \greygra{0.44} & \greygra{0.53} & \greygra{0.35} & \greygra{0.51} & \greygra{0.32} & \greygra{0.51} & \greygra{0.43} & \textbf{\greygra{0.52}} & \textbf{\greygra{0.38}} &  & \greygra{0.56} & \greygra{0.47} & \greygra{0.54} & \greygra{0.41} & \greygra{0.52} & \greygra{0.42} & \greygra{0.50} & \greygra{0.22} & \textbf{\greygra{0.53}} & \textbf{\greygra{0.38}} \\
\addlinespace[3pt]
 & Revise & 0.53 & 0.63 & 0.52 & 0.60 & 0.52 & 0.65 & 0.52 & 0.62 & \textbf{\greygra{0.52}} & \textbf{\greygra{0.62}} &  & 0.53 & 0.56 & 0.52 & 0.59 & 0.52 & 0.53 & 0.51 & 0.59 & \textbf{\greygra{0.52}} & \textbf{\greygra{0.56}} &  & 0.53 & 0.59 & 0.54 & 0.49 & 0.51 & 0.66 & 0.51 & 0.66 & \textbf{\greygra{0.52}} & \textbf{\greygra{0.60}} \\
 & GECScore & 0.65 & 0.64 & 0.62 & 0.59 & 0.64 & 0.62 & 0.62 & 0.60 & \textbf{\greygra{0.63}} & \textbf{\greygra{0.61}} &  & 0.66 & 0.63 & 0.62 & 0.59 & 0.61 & 0.60 & 0.55 & 0.54 & \textbf{\greygra{0.61}} & \textbf{\greygra{0.59}} &  & 0.63 & 0.57 & 0.57 & 0.47 & 0.56 & 0.48 & 0.50 & 0.00 & \textbf{\greygra{0.57}} & \textbf{\greygra{0.38}} \\
 & FDGPT (BB) & 0.53 & 0.55 & 0.51 & 0.49 & 0.50 & 0.01 & 0.53 & 0.52 & \textbf{\greygra{0.52}} & \textbf{\greygra{0.39}} &  & 0.55 & 0.57 & 0.53 & 0.52 & 0.51 & 0.22 & 0.52 & 0.32 & \textbf{\greygra{0.53}} & \textbf{\greygra{0.41}} &  & 0.57 & 0.53 & 0.53 & 0.46 & 0.52 & 0.51 & 0.50 & 0.00 & \textbf{\greygra{0.53}} & \textbf{\greygra{0.37}} \\
\cdashline{2-34} \addlinespace[1pt]
 & Avg. Black-box & \greygra{0.57} & \greygra{0.61} & \greygra{0.55} & \greygra{0.56} & \greygra{0.55} & \greygra{0.42} & \greygra{0.56} & \greygra{0.58} & \textbf{\greygra{0.56}} & \textbf{\greygra{0.54}} &  & \greygra{0.58} & \greygra{0.59} & \greygra{0.56} & \greygra{0.57} & \greygra{0.55} & \greygra{0.45} & \greygra{0.52} & \greygra{0.48} & \textbf{\greygra{0.55}} & \textbf{\greygra{0.52}} &  & \greygra{0.58} & \greygra{0.56} & \greygra{0.55} & \greygra{0.47} & \greygra{0.53} & \greygra{0.55} & \greygra{0.50} & \greygra{0.22} & \textbf{\greygra{0.54}} & \textbf{\greygra{0.45}} \\
\addlinespace[3pt]
 & xlm-RoBERTa & 0.64 & 0.59 & 0.59 & 0.57 & 0.59 & 0.58 & 0.64 & 0.61 & \textbf{\greygra{0.61}} & \textbf{\greygra{0.59}} &  & 0.63 & 0.60 & 0.66 & 0.65 & 0.58 & 0.58 & 0.68 & 0.67 & \textbf{\greygra{0.64}} & \textbf{\greygra{0.63}} &  & 0.63 & 0.62 & 0.68 & 0.68 & 0.54 & 0.52 & 0.52 & 0.45 & \textbf{\greygra{0.59}} & \textbf{\greygra{0.57}} \\
 & mDeBERTa & 0.62 & 0.58 & 0.60 & 0.60 & 0.58 & 0.58 & 0.62 & 0.62 & \textbf{\greygra{0.61}} & \textbf{\greygra{0.59}} &  & 0.64 & 0.58 & 0.68 & 0.68 & 0.66 & 0.66 & 0.69 & 0.69 & \textbf{\greygra{0.67}} & \textbf{\greygra{0.65}} &  & 0.62 & 0.58 & 0.69 & 0.69 & 0.61 & 0.60 & 0.64 & 0.64 & \textbf{\greygra{0.64}} & \textbf{\greygra{0.63}} \\
\cdashline{2-34} \addlinespace[1pt]
 & Avg. Supervised & \greygra{0.63} & \greygra{0.59} & \greygra{0.60} & \greygra{0.59} & \greygra{0.59} & \greygra{0.58} & \greygra{0.63} & \greygra{0.62} & \textbf{\greygra{0.61}} & \textbf{\greygra{0.59}} &  & \greygra{0.63} & \greygra{0.59} & \greygra{0.67} & \greygra{0.67} & \greygra{0.62} & \greygra{0.62} & \greygra{0.68} & \greygra{0.68} & \textbf{\greygra{0.65}} & \textbf{\greygra{0.64}} &  & \greygra{0.63} & \greygra{0.60} & \greygra{0.69} & \greygra{0.68} & \greygra{0.57} & \greygra{0.56} & \greygra{0.58} & \greygra{0.54} & \textbf{\greygra{0.62}} & \textbf{\greygra{0.60}} \\
\addlinespace[3pt]
\bottomrule
\end{tabular}
    
    \end{adjustbox}
    \caption{Detection accuracy (ACC) and F1-scores (F1) across tasks, languages, and models. Gray highlights average performances across detector families by generator (rows) and across generators by detector (bold columns).}
    \label{tab:main}
\end{table*}

\section{Experimental Setup}

In this section, we introduce generators, detectors, benchmark construction, and evaluation metrics. 

\paragraph{Generators} We generate MGT using four multilingual models from two families: proprietary and open-weight. We select models based on their ranking at the time of writing on LM Arena,\footnote{\url{https://blog.lmarena.ai/}} an open-source platform for crowdsourced AI benchmarking.  
For proprietary models, we use \textbf{GPT-4o mini}\footnote{\url{https://openai.com/index/gpt-4o-mini-advancing-cost-efficient-intelligence/}} and \textbf{Gemini 2.0 Flash}.\footnote{\url{https://deepmind.google/technologies/gemini/flash/}}  
For open-weight models, we select~\textbf{Qwen2.5-7B-Instruct}\footnote{\url{https://huggingface.co/Qwen/Qwen2.5-7B-Instruct}} and \textbf{Mistral-7B-Instruct}.\footnote{\url{https://huggingface.co/mistralai/Mistral-7B-Instruct-v0.3}}  
We opt for smaller models in this category to better align with our editor-driven writing task scenarios.

\paragraph{Detectors} We evaluate six detectors from three different families: training-based, zero-shot white-box, and zero-shot black-box methods. We consider only multilingual LLMs for all families.  
Specifically, we use~\textbf{XLM-RoBERTa}~\cite{conneau2020} and \textbf{mDeBERTa}~\cite{he2023} as training-based detectors, which we fine-tune with hyperparameter search; \textbf{Binoculars}~\cite{hans2024spotting},~\textbf{LLR}~\cite{su2023detectllm}, and~\textbf{FastDetectGPT (White-Box)}~\cite{hans2024spotting} as zero-shot white-box detectors; and \textbf{Revise-Detect}~\cite{zhu2023beat}, \textbf{GECScore}~\cite{wu2025who}, and \textbf{FastDetectGPT (Black-Box)}~\cite{hans2024spotting} as zero-shot black-box detectors.  
Appendix~\ref{app:detect} provides an overview and implementation details of each detector.

\paragraph{WETBench} We construct our benchmarking data by randomly sampling 2,700 HWT per task from the corresponding subsets of WikiPS and mWNC.  
For Paragraph Writing and Summarisation, we balance each subset by length tertiles; for TST, we evaluate at the sentence level for all languages and at the paragraph level for English only.  
For each task–language subset, we generate MGT using the four generators introduced above, applying the best-performing prompts from our prompt evaluation in Section~\ref{sec:task}: Naive RAG for Paragraph Writing, one-shot prompting for Summarisation, and five-shot prompting for TST.  
Our benchmark corpus comprises 101,940 human- and machine-written texts across tasks, languages, and generators. Appendix Table~\ref{tab:datastats} presents benchmark statistics.

\paragraph{Evaluation Metrics} Given the parallel nature of our benchmark data, our main evaluation metric is accuracy. We additionally report F1-scores, which represent the weighted harmonic mean of precision and recall.

\section{Results}
\label{sec:res}

Table~\ref{tab:main} presents our benchmarking results. Our main results are:  
(\textit{i}) our benchmark challenges detectors, which achieve considerably lower scores than in prior work~\cite[e.g.,][]{macko2023,guo2023closechatgpthumanexperts,li2023mage,wang2023m4,wang2024m4gt}, (\textit{ii}) supervised detectors significantly outperform zero-shot methods across all tasks and languages, and (\textit{iii}) detection accuracy is highest for summarisation, followed by slightly lower accuracy for paragraph writing, and lowest for TST. The following presents the most relevant trends by task.

\paragraph{Paragraph Writing} 

Across languages and models, training-based detectors outperform zero-shot methods by 19–30\% accuracy on average.  
Black-box detectors are 3–6\% more accurate than white-box detectors in English and Portuguese but perform slightly worse in Vietnamese.  
Only white-box detectors show a slight increase in accuracy when moving from high- to low-resource languages.

Within detector families, we do not find any substantial differences between the two training-based models.  
Among white-box detectors, Binoculars achieves up to 11\% higher accuracy, with performance on low-resource languages approaching that of training-based methods.  
For black-box detectors, GECScore exhibits up to 25\% higher accuracy compared to other models in its category.

Considering generators, training-based detectors achieve on average 2–7\% higher accuracy on smaller-sized generators.  
This pattern is reversed for zero-shot detectors, where accuracy is higher for larger models, with the gap widening in lower-resource languages.  
These results suggest that when generating paragraphs from scratch, smaller models leave more detectable semantic and syntactic traces for training-based detectors.  
In contrast, the internals and token-level patterns of larger models seem to exhibit stronger signals than those of smaller models for zero-shot methods.

We observe substantial anomalies in Mistral’s output for Vietnamese.  
The model often fails to follow prompts, producing unclear outputs that provide simple cues for training-based detectors but appear to confuse zero-shot methods.  
We provide an analysis of Mistral’s generation issues in Appendix~\ref{app:mistral}.

\paragraph{Summarisation} 

Among all tasks, supervised detectors perform best on summarisation, achieving an average accuracy of 89\% across languages and generators.  
An exception is Gemini 2.0, for which detection accuracies are on average 6–17\% lower, suggesting that its summaries may more closely resemble human-written references.  
While most LLMs are trained on summarisation tasks, enabling strong zero-shot performance~\cite{ouyang2022}, Wikipedia lead sections follow a distinctive style and formatting that seem to provide strong cues for training-based detectors.

Compared to Paragraph Writing, average detection accuracies for zero-shot models are slightly higher.  
White-box detectors achieve 4\% higher accuracy on English summaries compared to black-box models, while performance is similar for Portuguese and Vietnamese.  
As in Paragraph Writing, Binoculars achieves the highest average accuracy (65\%) among black-box detectors across languages and models, while GECScore performs best among white-box methods (68\%).

In contrast to Paragraph Writing, zero-shot metrics show little variation across generators for English.  
However, for Portuguese and Vietnamese, a similar pattern emerges: summaries generated by larger models are slightly easier to detect.  
This effect is less pronounced than in Paragraph Writing, with an average accuracy difference of around 6\%.

We attribute the similar trends between Paragraph Writing and Summarisation to the nature of both tasks: each involves generating text from scratch, conditioned either on retrieved context or article content.  
We observe the same issues as before for Mistral's Vietnamese summaries.
 
\paragraph{TST} 

For sentence-level TST, we observe the lowest accuracy scores across all tasks and detector families.  
While zero-shot detectors average between 52–56\% across languages and generators, training-based methods achieve only slightly higher scores, ranging from 61–65\%.  
A notable exception is GECScore, which outperforms other zero-shot methods by up to 12\%.

\begin{table}
    \centering
    
        \begin{adjustbox}{width=\linewidth}
    
        \begin{tabular}{lcccccccccc}
        \toprule
        \textbf{Detector} & \multicolumn{10}{c}{\textbf{TST English Paragraphs}} \\
        \cmidrule(lr){2-11}
         & \multicolumn{2}{c}{GPT-4o mini} & \multicolumn{2}{c}{Gemini 2.0} & \multicolumn{2}{c}{Qwen 2.5} & \multicolumn{2}{c}{Mistral} & \multicolumn{2}{c}{\textbf{Avg}} \\
        \cmidrule(lr){2-3} \cmidrule(lr){4-5} \cmidrule(lr){6-7} \cmidrule(lr){8-9} \cmidrule(lr){10-11}
         & ACC & F1 & ACC & F1 & ACC & F1 & ACC & F1 & ACC & F1 \\
        \midrule
        Binoculars & 0.58 & 0.53 & 0.55 & 0.47 & 0.52 & 0.39 & 0.57 & 0.51 & \textbf{\greygra{0.56}} & \textbf{\greygra{0.48}} \\
        LLR & 0.52 & 0.25 & 0.51 & 0.22 & 0.50 & 0.03 & 0.53 & 0.40 & \textbf{\greygra{0.51}} & \textbf{\greygra{0.22}} \\
        FDGPT (WB) & 0.60 & 0.63 & 0.56 & 0.60 & 0.52 & 0.60 & 0.58 & 0.61 & \textbf{\greygra{0.56}} & \textbf{\greygra{0.61}} \\
        \cdashline{1-11} \addlinespace[1pt]
        Avg (White-box) & \greygra{0.57} & \greygra{0.47} & \greygra{0.54} & \greygra{0.43} & \greygra{0.52} & \greygra{0.34} & \greygra{0.56} & \greygra{0.51} & \textbf{\greygra{0.55}} & \textbf{\greygra{0.44}} \\
        \addlinespace[3pt]
        Revise & 0.53 & 0.62 & 0.52 & 0.56 & 0.53 & 0.42 & 0.52 & 0.55 & \textbf{\greygra{0.53}} & \textbf{\greygra{0.54}} \\
        GECScore & 0.83 & 0.82 & 0.64 & 0.67 & 0.73 & 0.69 & 0.67 & 0.69 & \textbf{\greygra{0.72}} & \textbf{\greygra{0.72}} \\
        FDGPT (BB) & 0.59 & 0.62 & 0.54 & 0.55 & 0.51 & 0.63 & 0.58 & 0.54 & \textbf{\greygra{0.56}} & \textbf{\greygra{0.59}} \\
        \cdashline{1-11} \addlinespace[1pt]
        Avg (Black-box) & \greygra{0.65} & \greygra{0.69} & \greygra{0.57} & \greygra{0.59} & \greygra{0.59} & \greygra{0.58} & \greygra{0.59} & \greygra{0.59} & \textbf{\greygra{0.60}} & \textbf{\greygra{0.61}} \\
        \addlinespace[3pt]
        xlm-RoBERTa & 0.78 & 0.77 & 0.78 & 0.77 & 0.78 & 0.77 & 0.71 & 0.71 & \textbf{\greygra{0.76}} & \textbf{\greygra{0.76}} \\
        mDeBERTa & 0.83 & 0.83 & 0.77 & 0.76 & 0.81 & 0.81 & 0.67 & 0.64 & \textbf{\greygra{0.77}} & \textbf{\greygra{0.76}} \\
        \cdashline{1-11} \addlinespace[1pt]
        Avg (Supervised) & \greygra{0.81} & \greygra{0.80} & \greygra{0.78} & \greygra{0.77} & \greygra{0.80} & \greygra{0.79} & \greygra{0.69} & \greygra{0.67} & \textbf{\greygra{0.77}} & \textbf{\greygra{0.76}} \\
        \addlinespace[3pt]
        \bottomrule
        \end{tabular}
        \end{adjustbox}
    \caption{Detection accuracy (ACC) and F1-scores (F1) for TST English paragraphs. Gray highlights average performances across detector families by generator (rows) and across generators by detector (bold columns).}
    \label{tab:tst}
\end{table}

We attribute part of the reduced performance to the sentence-level setting.  
Comparing the English sentence-level results in Table~\ref{tab:main} to the paragraph-level results in Table~\ref{tab:tst}, we observe accuracy gains of up to 18\%, depending on the model.  
However, these improvements mostly apply to white-box and training-based models.

Compared to Paragraph Writing and Summarisation, TST involves only minimal modifications to human-written text.  
While detection scores on English paragraphs are slightly lower than for full generation from scratch, they remain substantially higher than for sentence-level TST.  
This suggests that training-based detectors can identify similarly strong MGT signals in paragraph-level text, regardless of whether the content is generated from scratch or modified at the token level.

\section{Conclusion}
\label{sec:conc}

We present \textbf{WETBench}, a multilingual, multi-generator benchmark for detecting MGT in task-specific Wikipedia editing scenarios.  
We build the benchmark from two new large-scale, multilingual Wikipedia text corpora—\textbf{WikiPS} and \textbf{mWNC}—which support a range of tasks relevant to the Wikipedia and AI communities.
Based on these data, we define three representative tasks, evaluate multiple prompting strategies, generate MGT from diverse LLMs using the best-performing prompts, and benchmark detectors.

Our benchmark reveals that detectors from diverse families underperform on our data, with substantial variation across languages, models, and tasks.  
Training-based detectors consistently outperform zero-shot methods but achieve only moderate detection accuracy.  
These results indicate that existing detectors struggle to generalise beyond generic setups, highlighting uncertainty around their reliability and effectiveness in real-world, editor-driven MGT scenarios on Wikipedia.

In future work, we plan to extend the benchmark with additional tasks, generators, and languages.  
We also aim to investigate the generalisability of our findings to open-ended generation tasks and other domains.

\section*{Limitations}

\paragraph{Editing Task Selection}  
We identify three common editing tasks, based on~\citet{ford2023wiki}, that vary in editing intensity. However, many other relevant editing tasks exist, reflecting different forms of content transformation.  
In particular, \textit{text translation} is a critical use case across many language editions of Wikipedia, as it helps bridge content gaps.  
Given the increasing capabilities of LLMs in translation~\cite{jiao2023chatgpt,zhu2023multilingual,yan2024gpt}, and the associated risks (see Section~\ref{sec:intro}), detecting machine-generated translations is an important and underexplored task.  
Similarly, there are alternative approaches to TST, such as grammar and spelling correction, which are highly relevant, especially for non-native Wikipedia editors.

\paragraph{Real-World Relevance of Editing Tasks}  
Our task selection is grounded in the study by~\citet{ford2023wiki}, which explores how Wikipedia editors perceive opportunities for AI-assisted writing.  
However, we acknowledge that our benchmark does not fully capture how MGT actually arises in real-world Wikipedia usage.  
While our tasks are motivated by plausible scenarios, we lack empirical evidence that editors systematically use LLMs in the ways we design them.  
Nonetheless, the findings of~\citet{ford2023wiki} provide the most systematic basis for aligning our benchmark with real-world editorial contexts.

\paragraph{NPOV Detection}  
To identify the most effective prompting technique for TST, we train four style classifiers per language-level setting (see Appendix Table~\ref{tab:sc}).  
However, our classifiers for Vietnamese and English at the paragraph level achieve accuracy only slightly above random chance, which might compromise the prompt evaluation in Section~\ref{sec:tst}.  
Despite extensive fine-tuning across model types, data, and hyperparameters, performance remains limited.  
For both subsets, we report the most conservative results to ensure that, even if classifier performance is poor, the precision of NPOV-related revisions is maximised (see Appendix~\ref{app:tst} for details).  
We acknowledge that NPOV detection on Vietnamese and English paragraph-level data is intrinsically challenging.

\paragraph{Text Length}  
When comparing detection results between sentence- and paragraph-level TST, we find that text length significantly affects performance.  
While we stratify samples by tertiles to control for length, we do not further analyse detection performance based on length, instead reporting average metrics.  
Given its impact, we plan to investigate text-length heterogeneity in future work.

\paragraph{Generalisability}  
Although we aim to cover a broad range of detectors, generators, and languages, our conclusions are limited to the evaluated settings.  
Due to the rapid pace of AI research, our configurations may quickly become outdated.
For example, through advances in LLMs or MGT detectors.  
To support ongoing progress, we open-source our data and benchmark and plan to maintain the repository to ensure its continued relevance.

\section*{Ethics Statement}
\label{sec:es}

Our work uses publicly available content from Wikipedia, licensed under CC BY-SA.  
We include no private or sensitive information, and our experiments pose no risk to Wikipedia editors or the Wikipedias under study.  
Sensitive data about individual contributors are neither identifiable nor exposed in any way.

We obtain machine-generated data using four LLMs under their respective licences:
\begin{itemize}
    \item GPT-4-mini: No specific license. OpenAI welcomes research publications.\footnote{\url{https://openai.com/policies/sharing-publication-policy/}}
    \item Gemini 2.0: Apache 2.0\footnote{\url{https://github.com/google-gemini}}
    \item QWen 2.0: Apache 2.0\footnote{\url{https://github.com/QwenLM/Qwen2.5}}
    \item Mistral: Apache 2.0\footnote{\url{https://mistral.ai/news/announcing-mistral-7b}}
\end{itemize}

This study addresses limitations in prior evaluations of SOTA MGT detectors by systematically assessing their performance in realistic editorial contexts.  
Our goal is to provide more accurate and practical insights into the feasibility and utility of MGT detection in collaborative knowledge environments such as Wikipedia.  
We emphasise that our experiments aim to inform the potential role of MGT detectors as automated metrics or as tools to assist users in identifying machine-generated content.

\section*{Acknowledgements}

This work was supported by the Engineering and Physical Sciences Research Council [grant number Y009800/1], through funding from Responsible AI UK (KP0011), as part of the Participatory Harm Auditing Workbenches and Methodologies (\href{https://phawm.org/}{PHAWM}) project.
Additional support was provided by UK Research and Innovation [grant number EP/S023356/1] through the UKRI Centre for Doctoral Training in Safe and Trusted Artificial Intelligence (\href{https://safeandtrustedai.org/}{www.safeandtrustedai.org}).

\bibliography{custom}

\clearpage
\appendix

\begin{table*}
    \centering

    \begin{adjustbox}{width=\linewidth}
        \begin{tabular}{ccccccccc}
            \hline
            \textbf{Corpus} & \textbf{Subset} & \textbf{Level} & \textbf{Language} & \textbf{Corpus N} & \textbf{Processed Corpus N} & \textbf{Eval N} & \textbf{Experiment N} & \textbf{MGT N} \\
            \hline

            \multirow{4}{*}{\textbf{extendWNC}} 
                & \multirow{4}{*}{Text Style Transfer} 
                & \multirow{3}{*}{Sentences}  & EN & 2,333,143 & 286,626 & 270 & 2,700 & 10,800 \\ 
                &                             &                             & PT & 31,506 & 7,877 &270 & 2700 & 10,800 \\ 
                &                             &                             & VI & 13,800 & 1,185 & 270 & 1185 & 4,740 \\ \cline{3-9}
                &                             & Paragraphs                  & EN & 4,671 & 4,671 & 270 & 2700 & 10,800 \\ 
            \hline

            \multirow{6}{*}{\textbf{WikiPS}} 
                & \multirow{3}{*}{Paragraph Writing}  
                &     & EN & 96,860 & 96,860 & 270 & 2700 & 10,800 \\
                &                                     &                         & PT & 72,965 & 72,965 & 270 & 2700 & 10,800 \\ 
                &                                     &                         & VI & 98,315 & 98,315 & 270 & 2700 & 10,800 \\ \cline{3-9}
                & \multirow{3}{*}{Summarisation}  
                &                                     & EN & 67,267 & 53,203 & 270 & 2700 & 10,800 \\ 
                &                                     &                         & PT & 56,538 & 36,075 & 270 & 2700 & 10,800 \\  
                &                                     &                         & VI & 60,884 & 45,500 & 270 & 2700 & 10,800 \\
            \hline
            \textbf{Total} &  & & & & & \textbf{2,700} & \textbf{25,485} & \textbf{101,940} \\
            \hline
        \end{tabular}
    \end{adjustbox}
    \caption{WETBench Dataset Statistics. Corpus N denotes the raw number of observations; Processed N denotes the number of observations after processing; Experiment N denotes the number of human-written texts; and MGT N denotes the total number of machine-generated texts.}
    \label{tab:datastats}
\end{table*}

\section{Data Construction}
\label{app:cons}

We download the meta stub history WikiDumps\footnote{\url{https://dumps.wikimedia.org/}} for all three languages, which serve as the foundational datasets for both \textbf{WikiPS} and the \textbf{mWNC}.  
For both datasets, we consider only the most recent instances---revisions for mWNC and article versions for WikiPS---that occurred prior to the public release of ChatGPT on 30 November 2022.  
This filtering step ensures that our data is not contaminated by MGT.

\subsection{WikiPS}

We begin by retrieving the latest revision IDs for all \textit{articles} (excluding discussion pages and other non-content pages) in each target language.  
We then randomly sample and crawl these articles by querying the MediaWiki Action API\footnote{\url{https://www.mediawiki.org/wiki/API:Main_page}\label{wikiapi}} until we collect 100{,}000 non-stub Wikipedia articles in HTML format per language.  
Rather than concentrating on a set of topics, we rely on a large enough random sample to provide a representative snapshot of each Wikipedia.  
We also rely on HTML representations, as parsing raw MediaWiki markup often leads to errors and occasional information loss (e.g., incomplete internal links).

We filter out articles lacking essential structural elements, such as a title, lead section, content sections, or references, as well as list-based articles.  
From the remaining articles, we use \texttt{BeautifulSoup} to pre-process, clean, and parse the HTML and extract the following components: the lead section, infobox (if available), paragraphs with their section headers (excluding sections such as ``See also'', ``External links'', etc.), and reference lists.  
This process yields article-level corpora of 67{,}267 articles in English, 56{,}538 in Portuguese, and 60{,}884 in Vietnamese.

\paragraph{Paragraphs} To construct our paragraph-level dataset, we randomly sample 20{,}000 articles per language.  
We then define a paragraph as a block of text containing at least three sentences and a minimum of 20 characters.  
For each paragraph, we collect metadata including its position within the article and any associated external references.  
We further refine the dataset by removing paragraphs without any references and those whose token counts fall outside two standard deviations from the mean token count of the corpus.  
Based on the filtered corpus, we compute tertiles for each paragraph and assign each to its corresponding range (EN $(83.0, 120.0)$; PT $(88.0, 128.0)$; VI $(108.0, 160.0)$).  
For the Paragraph Writing task, we only consider the first paragraph following a section or subsection.  
The resulting raw paragraph-level corpora consist of 96{,}860 paragraphs in English, 72{,}965 in Portuguese, and 98{,}315 in Vietnamese.

\paragraph{Summaries} To construct our summarisation dataset, we extract the lead section, infobox, and article body from the processed text corpora for each article.  
For the English and Portuguese corpora, we exclude lead sections with fewer than 10 tokens or with token lengths exceeding two standard deviations above the token mean.  
Similarly, we discard article bodies with fewer than 100 tokens or more than two standard deviations above the mean token count.  
For the Vietnamese corpus, whose article bodies are considerably longer (see Table~\ref{tab:sumcomp}), we set an upper limit of either 2,900 tokens or two standard deviations above the mean to mitigate context length constraints during model processing.  
As we treat each component as text input, we apply minimal markdown-like formatting to both the infobox and article body, such as rendering headers in bold.  
The resulting summarisation corpora consist of 53{,}203 lead–article pairs in English, 36{,}075 in Portuguese, and 45{,}500 in Vietnamese.

\begin{table}
    \centering
    \footnotesize
    \begin{adjustbox}{width=\linewidth}

    % INSERT TABLE HERE
    \begin{tabular}{lcccccc}
    \toprule
    Metric/Corpus & \textbf{WikiLingua} & \textbf{CNN/DM} & \textbf{arXiv} & \textbf{WikiSums EN} & \textbf{WikiSums PT} & \textbf{WikiSums VI} \\
    \midrule
    \textbf{Size} & 142,346 & 311,971 & 215,913 & 67,267 & 56,538 & 60,884 \\
    \textbf{Summary Length} & 32 (19) & 51 (21) & 272 (572) & 83 (78) & 87 (95) & 135 (148) \\
    \textbf{Body Length} & 379 (224) & 690 (337) & 6029 (4570) & 667 (1027) & 587 (1121) & 940 (1800) \\
    \textbf{Infobox Length} & - & - & - & 61 (60) & 62 (40) & 95 (78) \\
    \textbf{ROUGE-1} & 0.13 & 0.14 & 0.07 & 0.17 & 0.18 & 0.30 \\
    \textbf{ROUGE-2} & 0.05 & 0.08 & 0.04 & 0.06 & 0.06 & 0.16 \\
    \textbf{Compression Rate} & 14.12 & 14.66 & 39.78 & 10.30 & 7.80 & 8.56 \\
    \textbf{Novel Unigram \%} & 0.38 & 0.20 & 0.15 & 0.53 & 0.62 & 0.49 \\
    \textbf{Novel Bigram \%} & 0.78 & 0.60 & 0.45 & 0.83 & 0.88 & 0.80 \\
    \textbf{Novel Trigram \%} & 0.93 & 0.77 & 0.69 & 0.93 & 0.95 & 0.91 \\
    \midrule
    \textbf{Entity Sample Size} & 20,000 & 20,000 & 20,000 & 20,000 & 20,000 & 20,000 \\
    \textbf{Entity F1-Score} & 0.06 & 0.21 & 0.04 & 0.14 & 0.17 & 0.13 \\
    \bottomrule
    \end{tabular}

    \end{adjustbox}
    \caption{Summarisation Corpora Comparison. Numbers in parentheses report standard deviation.}
    \label{tab:sumcomp}
    
\end{table}

Table~\ref{tab:sumcomp} compares our raw summarisation datasets to three commonly used benchmarks from different domains: WikiLingua~\cite{ladhak2020} for Wikimedia content, CNN/DM~\cite{nallapati2016} for news, and arXiv~\cite{cohan2018} for academic writing.

On average, our summaries are considerably longer than those in WikiLingua and CNN/DM, but shorter than arXiv abstracts.  
The average body length in our datasets is comparable to CNN/DM but significantly shorter than arXiv.  
Despite this, our datasets exhibit higher ROUGE-1 and ROUGE-2 scores, indicating improved content overlap.  
We also observe lower compression rates~\cite{grusky2018}, meaning our summaries are proportionally longer relative to article bodies.  
Furthermore, our datasets show consistently higher percentages of novel unigrams, bigrams, and trigrams, suggesting a greater degree of abstractiveness.

To address the concern that a higher proportion of novel tokens may signal information asymmetry between the lead and the article body, we compute entity overlap F1-scores on a 20,000-example subset of each dataset.  
Our results show higher entity F1-scores compared to WikiLingua, CNN/DM, and arXiv, indicating that our datasets maintain a comparable or better level of factual consistency.

Among the Wikipedias, the Vietnamese edition features leads, infoboxes, and article bodies that are approximately 30\% longer than their English and Portuguese counterparts.  
Despite higher ROUGE scores, the comparable share of novel n-grams in Vietnamese indicates a slightly lower level of abstractiveness relative to the other language versions.

\subsection{mWNC}

We largely follow the procedure of~\citet{pryzant2020}, with modifications to accommodate larger multilingual datasets.  
From each Wikidump, we extract all NPOV-related revisions made prior to the release of ChatGPT.  
We expand the set of NPOV-related keywords (e.g., NPOV, POV, neutral, etc.) for each Wikipedia edition based on its respective NPOV policy page.\footnote{English: \href{https://en.wikipedia.org/wiki/Wikipedia:Neutral_point_of_view}{Neutral point of view}; Portuguese: \href{https://pt.wikipedia.org/wiki/Wikip\%C3\%A9dia:Princ\%C3\%ADpio_da_imparcialidade}{Princípio da imparcialidade}; Vietnamese: \href{https://vi.wikipedia.org/wiki/Wikipedia:Th\%C3\%A1i_\%C4\%91\%E1\%BB\%99_trung_l\%E1\%BA\%ADp}{Thái độ trung lập}}  
This yields 2,333,143 relevant revisions for English, 31,506 for Portuguese, and 13,800 for Vietnamese.

We retrieve the corresponding diffs\footnote{\url{https://en.wikipedia.org/wiki/Help:Diff}} using the MediaWiki API,\footnote{\footref{wikiapi}} which we extensively clean and pre-process.  
To match pre- and post-neutralisation sentence pairs within each edit chunk, we first discard all unedited sentences and then apply pairwise BLEU scoring to identify the highest-scoring sentence pairs.  
For details on chunk and sentence filtering, we refer to~\citet{pryzant2020}.

Our main modifications include: (\textit{1}) retaining reverts, and (\textit{2}) for Vietnamese only, relaxing the Levenshtein distance threshold to $<$3 and allowing up to two edit chunk pairs and multiple sentence-level matches.  
This adjustment addresses the comparatively low number of NPOV-related edits in Vietnamese, which would otherwise yield only a few hundred usable instances.

These modifications result in 286,626 sentence pairs for English, 7,877 for Portuguese, and 1,185 for Vietnamese.  
While we could further increase $N$ for Vietnamese by loosening the filtering criteria, we find that this introduces noise and does not improve the performance of the downstream style classifier.  
We therefore prioritise a smaller, higher-precision dataset (see also Appendix~\ref{app:tst}).

Due to the stark disparity in data size, we obtain paragraph-level data only for English.  
For this, we construct a dataset that, like the Vietnamese setup, allows multiple edit chunk and sentence-level matches.  
We define a paragraph-level pair as one in which at least one addition or deletion occurs in each of three adjacent sentences.  
This yields a dataset of 4,671 paragraph pairs.

\section{Task Design Details}
\label{app:tasks}

For brevity, we present prompts in English only.

\subsection{Paragraph Writing}

\subsubsection{Paragraph Writing Prompts}

\textbf{Minimal}
\begin{lstlisting}
Please write the first paragraph for the section "{section_title}" in the Wikipedia article "{page_title}" using no more than {n_words} words. Only return the paragraph.
\end{lstlisting}

\noindent
\textbf{Content Prompts}
\begin{lstlisting}
Please write the first paragraph for the section "{section_title}" in the Wikipedia article "{page_title}". 

Address the following key points in your response:
{content_prompts}

Use no more than {n_words} words. Only return the paragraph.
\end{lstlisting}

\noindent
\textbf{RAG}
\begin{lstlisting}
Use the following context to ensure factual accuracy when writing:
{context}

--

Please write the first paragraph for the section "{section_title}" in the Wikipedia article "{page_title}".

Address the following key points in your response:
{content_prompts}

Use the context above to inform your response, in addition to any relevant knowledge you have. Use no more than {n_words} words. Only return the paragraph in {language}.
\end{lstlisting}

\subsubsection{Content Prompts}

We model editors' LLM-assisted content generation through Content Prompts. For instance, an editor aiming to expand a Wikipedia article might prompt a model to generate a paragraph in response to factual questions about a specific topic (e.g., "What are London’s most notable modern buildings?" or "What is London's tallest skyscraper?"), within a given section (e.g., Architecture).  
For each human-written paragraph in our dataset, we prompt GPT-4o~\cite{openai2024gpt4technicalreport} to generate a minimum of five content prompts for low-tertile paragraphs, and eight for medium- and high-tertile paragraphs.  
Although this method does not exhaustively cover all factual content from the HWT, it substantially improves the alignment of factual information between HWT and MGT.

\subsubsection{Naive RAG}
\label{app:rag}
We implement a web-based Naive RAG setup to reflect an editing scenario in which an editor, in addition to providing task instructions and content prompts, also supplies relevant context to minimise factual inaccuracies.  
Our RAG pipeline follows the indexing, retrieval, and generation modules of the Naive variant~\cite{gao2024rag}, with two key modifications: we prepend the pipeline with Content Prompts and Web Search modules.

\paragraph{Content Prompts and Web Search} For each paragraph, we generate diverse content prompts as described above.  
We use each content prompt to query the Google Custom Search API,\footnote{\url{https://developers.google.com/custom-search/v1/overview}} retrieving the top 10 most relevant URLs.  
From the search results, we exclude the original Wikipedia page (if applicable) as well as any unreliable sources~\cite{shao2024}.

\paragraph{Indexing} We download the raw HTML of each scrappable web page and apply a series of preprocessing and cleaning steps.  
We then split each page into chunks using LangChain’s RecursiveCharacterTextSplitter.\footnote{\href{https://api.python.langchain.com/en/latest/character/langchain_text_splitters.character.RecursiveCharacterTextSplitter.html}{LangChain RecursiveCharacterTextSplitter documentation}}  
We compute BGE-M3\footnote{\url{https://huggingface.co/BAAI/bge-m3}} embeddings for each chunk and store them in a vector database.

\paragraph{Retrieval and Generation} We treat each content prompt as a query, compute its embedding, and retrieve the two most similar chunks from the vector database based on cosine similarity.  
We append these retrieved chunks to the content prompt as context to guide the model’s generation.

\subsection{Summarisation}
\label{app:sum} % keep this!

\subsubsection{Prompts}

\textbf{Minimal}
\begin{lstlisting}
Your task is to summarize the below article with no more than {n_toks_trgt} words. Article:

"""{src}"""
\end{lstlisting}

\noindent
\textbf{Instruction/Few-Shot}
\begin{lstlisting}
Your task is to summarize an article to create a Wikipedia lead section. 
- In Wikipedia, the lead section is an introduction to an article and a summary of its most important contents. 
- Apart from basic facts, significant information should not appear in the lead if it is not covered in the remainder of the article.

Generate the lead for the article titled "{page_title}" using the article's body above with no more than {n_toks_trgt} words. Article:

"""{src}"""
\end{lstlisting}

\subsection{TST}
\label{app:tst}

\subsubsection{Prompts}

\textbf{Minimal}
\begin{lstlisting}
Please make this sentence/paragraph more neutral. **Make as few changes as possible and use no more than {trgt_n_words} words for the neutralised sentence/paragraph.** Sentence/Paragraph:

"""{src}"""
\end{lstlisting}

\noindent
\textbf{Instruction/Few-Shot}
\begin{lstlisting}
Please edit this biased Wikipedia sentence/paragraph to make it more neutral, aligning with Wikipedia's neutral point of view policy:

Achieving what the Wikipedia community understands as neutrality means carefully and critically analyzing a variety of reliable sources and then attempting to convey to the reader the information contained in them fairly, proportionately, and as far as possible without editorial bias. Wikipedia aims to describe disputes, but not engage in them. The aim is to inform, not influence. Editors, while naturally having their own points of view, should strive in good faith to provide complete information and not to promote one particular point of view over another. The neutral point of view does not mean the exclusion of certain points of view; rather, it means including all verifiable points of view which have sufficient due weight. Observe the following principles to help achieve the level of neutrality that is appropriate for an encyclopedia:

- Avoid stating opinions as facts.
- Avoid stating seriously contested assertions as facts.
- Avoid stating facts as opinions.
- Prefer nonjudgmental language.
- Do not editorialize.
- Indicate the relative prominence of opposing views.

**Make as few changes as possible and use no more than {trgt_n_words} words for the neutralised sentence/paragraph.** Output only the neutralized sentence/paragraph. Sentence/Paragraph:

"""{src}"""
\end{lstlisting}

\subsubsection{Style Classifiers}
\label{app:sc}

We fine-tune four style classifiers: one for each language at the sentence level, and an additional classifier for English at the paragraph level.  
The hyperparameter settings are provided in Table~\ref{tab:hyper}.

\begin{table}[!htbp]
    \centering
    \begin{adjustbox}{scale=0.5}

    \begin{tabular}{lccccc}
    \hline
    \textbf{Language/Level} & \textbf{Models} & \textbf{Learning Rate} & \textbf{Batch Sizes} & \textbf{Epochs} & \textbf{Weight Decay} \\
    \hline
    EN/Sent. & \makecell{roberta-base} & 1e-6 & 32 & 15 & 0.01 \\
    PT/Sent. & \makecell{xlm-roberta-base,\\mBERT} & 5e-5, 1e-5, 5e-6 & 16, 32 & 2, 5, 8 & 0, 0.01 \\
    VI/Sent. & \makecell{xlm-roberta-base,\\mBERT} & \makecell{5e-5, 1e-5,\\5e-6, 1e-6} & 16, 32 & 2, 4, 6 & 0, 0.01 \\ \hline 
    EN/Para. & \makecell{roberta-base} & \makecell{5e-5, 1e-6,\\5e-6} & 16, 32 & 3, 6, 9 & 0, 0.01 \\
    \hline
    \end{tabular}
    \end{adjustbox}

        \caption{Style Classifier Hyperparameter Settings.}
    \label{tab:hyper}
\end{table}

For English, we adopt the hyperparameters from the best-performing neutrality classifier available on Hugging Face.\footnote{\url{https://huggingface.co/cffl/bert-base-styleclassification-subjective-neutral}}  
As the English data contain nearly a quarter of a million sentence pairs, we fine-tune on a smaller subset of the most recent 150k pairs, specifically filtered to include the keyword \textit{NPOV} in the revision content, in order to further enhance precision.  
For Portuguese, we apply commonly used hyperparameter values, while for Vietnamese and English paragraphs, we extend the search space, as initial experiments yielded low detection performance.

\begin{table}[!htbp]
    \centering
    \begin{adjustbox}{width=\linewidth}
        
    \begin{tabular}{llccc}
        \toprule
        \textbf{Level} & \textbf{Language} & \textbf{Pairs} &  \textbf{Test Accuracy} \\
        \midrule
        \multirow{3}{*}{Sentences} & English & 300,000 & 73\% \\
                                    & Portuguese  & 5738  & 63\%\\
                                    & Vietnamese  & 2370  & 58\%\\ 
        \midrule
        Paragraphs & English & 9342 & 58\%\\
        \bottomrule
    \end{tabular}
    \end{adjustbox}
    \caption{Style Transfer Classifier Performance. Pairs denote biased and neutralised samples.}
    \label{tab:sc}

\end{table}

Table~\ref{tab:sc} reports the style classifier hyperparameter fine-tuning results.  
While fine-tuned models for English and Portuguese sentences yield satisfactory results, style accuracy for English paragraphs and Vietnamese sentences is low.  
In the following, we provide a qualitative analysis of both subsets and explain how we address these low performances.

\paragraph{Low Style Classifier Performance Analysis}

Table~\ref{tab:ex} presents two representative examples of NPOV revisions from each subset.  
The first example in each case illustrates a clear NPOV violation.  
For instance, the phrase "considered the best footballer" in Vietnamese and "not as strong" in English are both subjective.  
However, as illustrated with the second examples, NPOV filtering also captures revisions related to political or historical content, which often rely on (subjectively) factual corrections rather than systematic semantic cues.

\begin{table}[!h]
    \centering
    \footnotesize

    \begin{tabular}{cp{0.7\linewidth}}
    \toprule
    \textbf{Subset} & \textbf{Biased Examples} \\
    \hline
    \multirow{2}[50]{*}{Vietnamese} & \textit{\textbf{Được coi là cầu thủ xuất sắc nhất thế giới và là cầu thủ vĩ đại nhất mọi thời đại (Greatest of All Time - GOAT)}, Ronaldo là chủ nhân của 5 Quả bóng vàng châu Âu vào các năm 2008, 2013, 2014, 2016, 2017 và cũng là chủ nhân 4 Chiếc giày vàng châu Âu, cả hai đều là kỷ lục của một cầu thủ châu Âu cùng nhiều danh hiệu cao quý khác.} (EN: Considered the best football player in the world and the greatest of all time (GOAT), Ronaldo has won 5 Ballon d'Or awards in the years 2008, 2013, 2014, 2016, and 2017, as well as 4 European Golden Shoes—both records for a European player—along with many other prestigious titles.) \\
    \cmidrule{2-2}
                              & \textit{Ông từng phục vụ Lý Hoài Tiên, tướng dưới quyền \textbf{nghịch tặc} Sử Tư Minh của Ngụy Yên.} (EN: He once served Lý Hoài Tiên, a general under the command of the rebel Sử Tư Minh of Ngụy Yên.) \\
    \midrule
    \multirow{2}[50]{*}{\makecell{English \\ Paragraphs}} & \textit{He is \textbf{not as strong}, although still an exceptional warrior. Agamemnon clearly has a stubborn streak that \textbf{one can argue makes him even more arrogant} than Achilles. Although he takes few risks in battle, Agamemnon \textbf{still accomplishes great progress} for the Greeks.} \\
    \cmidrule{2-2}
                              & \textit{The population of Bangladesh ranks seventh in the world, but its area of approximately is ranked \textbf{ninety-fourth}, making it one of the most densely populated countries in the world, or the most densely populated country if small island nations and city-states are not included. It is the third-largest Muslim-majority nation, \textbf{but} has a smaller Muslim population than the Muslim minority in India. Geographically \textbf{dominated} by the fertile Ganges-Brahmaputra Delta, the country has annual monsoon floods, and cyclones are frequent.}\\
    \bottomrule
    \end{tabular}

    \caption{NPOV Revision Examples. Parentheses contain English translations. Highlighted words indicate words that were edited.}
    \label{tab:ex}

\end{table}

As we observed this pattern consistently across both subsets, we conducted additional data processing and hyperparameter tuning for the classifiers.  
We explored several strategies, including: (1) extending the list of NPOV-related keywords, (2) allowing multiple edit chunks per revision, (3) permitting multi-sentence edits within a single chunk, and (4) expanding the range of hyperparameter settings and model types.  
However, none of these approaches significantly improved style classifier performance.

Therefore, we selected the configuration that yielded the highest precision, adopting a conservative approach to extract NPOV-relevant revision pairs.  
Despite the relatively low classifier accuracy, we are confident that our dataset includes a high proportion of true positives.

\section{Detector Details and Implementations}
\label{app:detect}

We follow the taxonomy for detecting MGT proposed by~\cite{yang2023survey}, which categorises detectors into three types: 1) zero-shot, 2) training-based, and 3) watermarking, although we exclude the latter from our experiments.  
The taxonomy further divides zero-shot methods into white-box and black-box, depending on whether the detector has access to the generator's logits or other model internals.  
For all detectors, when the original baseline LLM does not support one of our languages, we replace it with a multilingual model of comparable size. For zero-shot detectors, we use Youden’s J statistic to determine the optimal threshold.

\subsection*{Zero-shot White-box}

\noindent
\textbf{LLR}~\cite{su2023detectllm} The Log-Likelihood Log-Rank Ratio (LLR) intuitively leverages the ratio of absolute confidence through log-likelihood to relative confidence through log rank for a given sequence.  
We implement this detector with Bloom-3B.\footnote{\url{https://huggingface.co/bigscience/bloom-3b}\label{ft:bloom}} \\

\noindent
\textbf{Binoculars}~\cite{hans2024spotting} Binoculars introduces a metric based on the ratio of perplexity to cross-perplexity, where the latter measures how surprising the next-token predictions of one model are to another.  
We implement this detector using Qwen2.5-7B\footnote{\url{https://huggingface.co/Qwen/Qwen2.5-7B}} for the observer model and Qwen2.5-7B-Instruct\footnote{\url{https://huggingface.co/Qwen/Qwen2.5-7B-Instruct}} for the performer model. \\

\noindent
\textbf{FastDetectGPT White-Box}~\cite{bao2023} DetectGPT~\cite{mitchell2023} exploits the observation that MGT tends to be located in regions of negative curvature in the log-probability function, from which a curvature-based detection criterion is defined.  
FastDetectGPT (WB) is an optimised version of DetectGPT that builds on the \textit{conditional} probability curvature.  
We implement the white-box version with Bloom-3B.\footref{ft:bloom}

\subsection*{Zero-shot Black-box}

\noindent
\textbf{Revise}~\cite{zhu2023beat} Revise builds on the hypothesis that ChatGPT\footnote{\href{https://openai.com}{https://openai.com}} performs fewer revisions when generating MGT, and thus bases its detection criterion on the similarity between the original and revised articles.  
We implement this detector as in the original paper, using GPT-3.5-turbo.\footnote{\url{https://platform.openai.com/docs/models/gpt-3.5-turbo}\label{gpt35}} \\

\noindent
\textbf{GECScore}~\cite{wu2025who} Grammar Error Correction Score assumes that HWT contain more grammatical errors and calculates a Grammatical Error Correction score.  
We implement this detector as in the original paper, using GPT-3.5-turbo.\footref{gpt35} \\

\noindent
\textbf{FastDetectGPT Black-Box}~\cite{hans2024spotting} In the black-box version, the scoring model differs from the reference model.  
We use BLOOM-3B as the reference model and BLOOM-1.7B as the scoring model.

\subsection*{Supervised}

\noindent
\textbf{XLM-RoBERTa}~\cite{conneau2020}: XLM-RoBERTa\footnote{\url{https://huggingface.co/FacebookAI/xlm-roberta-base}} is the multilingual version of RoBERTa~\cite{liu2019} for 100 languages. RoBERTa improves upon BERT~\cite{devlin2019} through longer and more extensive training, as well as dynamic masking. \\

\noindent
\textbf{mDeBERTaV3} mDeBERTaV3\footnote{\url{https://huggingface.co/microsoft/mdeberta-v3-base}} is the multilingual version of DeBERTa~\cite{he2023}, which enhances BERT and RoBERTa using disentangled attention and an improved masked decoder. \\

\begin{table}
    \centering
    \footnotesize
    \begin{tabular}{lc}
        \toprule
        \textbf{Hyperparameter} & \textbf{Values} \\
        \midrule
        Batch Size & 16, 32 \\
        Learning Rate & 1e-5, 5e-6, 1e-6 \\
        Epochs & 3, 5 \\
        \bottomrule
    \end{tabular}
    \caption{Hyperparameter settings for supervised-detectors.}
    \label{tab:exphp}
\end{table}

\noindent
Both models are fine-tuned per task and language on an 80/10/10 split with the hyperparameter choices displayed in Table~\ref{tab:exphp}.

\section{Additional Results}

\subsection{Mistral Error Analysis}
\label{app:mistral}

We observe anomalous evaluation metrics for Vietnamese texts written by Mistral. While both zero-shot detectors achieve random chance accuracy and often zero F1-scores, training-based detectors achieve near-perfect metrics. Upon inspecting the data, we find that Mistral, unlike the other models, fails to follow the instructions in our prompts. Common errors include outputting text mid-sentence or returning English text, despite the final sentences of our prompts emphasising that the response should be in Vietnamese. These flaws explain the strong performance of training-based detectors, as they detect such syntactic imperfections, whereas zero-shot detectors appear unable to identify clear patterns based on model internals or token-level features.

\section{Paper Checklist}

\subsection*{Benefits}

\paragraph{Q1} \textit{How does this work support the Wikimedia community?}

\paragraph{A1} We believe our work supports the Wikimedia community in at least two ways. First, we introduce two new text corpora that extend beyond MGT detection and can be leveraged for various AI applications. The mWNC dataset addresses (1) community requests to expand existing resources with additional languages, and (2) high-priority needs identified by workshop organizers for NPOV datasets to train and evaluate models for biased language detection. These data open up several research directions, such as training models to detect bias in longer text sequences and testing their generalisability across varying text lengths. As our style classifier results suggest, NPOV detection in low-resource languages remains challenging, making mWNC a valuable resource for advancing this area of research.

Likewise, with WikiPS, we aim to provide two large-scale subsets of general interest to the research community. The paragraph-level subset, for instance, can be used to build question answering datasets for non-high-resource languages, analogous to SQuAD~\cite{rajpurkar2016}. Our summarisation subset naturally lends itself to improving lead section summarisation models. We highlight the inclusion of infoboxes as a key input feature for lead generation, in line with recent findings that LLM-generated summaries are often on par with—or even preferred over—human-written ones~\cite{goyal2022news,pu2023summarization,zhang2024benchmarking}.

Second, our benchmark, WETBench, is designed to inform the Wikipedia community about the feasibility and effectiveness of current state-of-the-art detectors in identifying MGT instances on the platform. As outlined in the introduction, there is growing concern about the influx of low-quality, unreliable machine-generated content. Due to limitations in prior evaluations (see Section~\ref{sec:intro}), we hope our work contributes to a better understanding of the capabilities and limitations of current detectors, supporting future research and real-world efforts to identify and manage MGT on Wikipedia.

\paragraph{Q2} What license are you using for your data, code, models? Are they available for community re-use? 

\paragraph{A2} We release our datasets, WikiPS and mWNC, which are derived from Wikipedia, under the CC BY-SA 4.0 license. Users of the MGT included in our benchmark must ensure compliance with the respective licenses of each language model (see Ethics Statement). We open-source all code used in our work.

\paragraph{Q3} Did you provide clear descriptions and rationale for any filtering that you applied to your data? For example, did you filter to just one language (e.g., English Wikipedia) or many? Did you filter to any specific geographies or topics? \\

\paragraph{A3} We provide comprehensive explanations of our dataset construction in Section~\ref{sec:data} and Appendix~\ref{app:cons}. Section ~\ref{sec:data} outlines the high-level construction process and key design choices, while Appendix~\ref{app:cons} offers a detailed walkthrough for readers interested in replicating or closely examining our methodology. For fine-grained construction details, we refer readers to our publicly available codebase.

\subsection*{Risks}

\paragraph{Q1} If there are risks from your work, do any of them apply specifically to Wikimedia editors or the projects?

\paragraph{A1} Our research objective is to provide a more accurate assessment of SOTA MGT detectors' performance on task-specific MGT. We acknowledge that our findings could be misinterpreted or misused to claim that SOTA detectors are ineffective at identifying machine-assisted edits. However, the intent of our work is not to undermine the potential of detection methods but to highlight their current limitations in realistic editorial settings.

\paragraph{Q2} Did you name any Wikimedia editors (including username) or provide information exposing an editor's identity?

\paragraph{A2} No. Our data includes only textual information, without any references to individual editors.

\paragraph{Q3} Could your research be used to infer sensitive data about individual editors? If so, please explain further.

\paragraph{A3} No. While our dataset includes revision IDs, it does not contain any additional information that is not already publicly available on Wikipedia.

\end{document}